\documentclass{article} 
\usepackage{iclr2027_conference,times}

\usepackage{amsmath,amsfonts,bm}

\def\eqref#1{equation~\ref{#1}}

\def\1{\bm{1}}

\DeclareMathAlphabet{\mathsfit}{\encodingdefault}{\sfdefault}{m}{sl}
\SetMathAlphabet{\mathsfit}{bold}{\encodingdefault}{\sfdefault}{bx}{n}

\usepackage{booktabs}
\usepackage{caption}
\usepackage{float}
\usepackage{hyperref}
\usepackage{url}
\usepackage[most]{tcolorbox}
\usepackage{graphicx}
\usepackage{xcolor}

\definecolor{promptblue}{RGB}{43,60,165}
\definecolor{promptback}{RGB}{248,249,255}

\newtcolorbox{promptbox}[1]{
    enhanced,
    breakable,
    colback=promptback,
    colframe=promptblue,
    colbacktitle=promptblue,
    coltitle=white,
    title={#1},
    fonttitle=\bfseries,
    boxrule=0.8pt,
    arc=3mm,
    outer arc=3mm,
    left=3mm,
    right=3mm,
    top=2.5mm,
    bottom=2.5mm,
    titlerule=0pt,
    toptitle=1.5mm,
    bottomtitle=1.5mm
}
\usepackage{listings}
\usepackage{xcolor}

\lstdefinestyle{compactpython}{
    language=Python,
    basicstyle=\ttfamily\fontsize{8.0}{8.6}\selectfont,
    keywordstyle=\color{magenta},
    commentstyle=\color{gray},
    stringstyle=\color{teal!70!black},
    numberstyle=\tiny,
    frame=single,
    rulecolor=\color{gray!35},
    backgroundcolor=\color{gray!2},
    showstringspaces=false,
    breaklines=true,
    breakatwhitespace=false,
    columns=fullflexible,
    keepspaces=true,
    tabsize=4,
    xleftmargin=3pt,
    xrightmargin=3pt,
    framexleftmargin=3pt,
    framexrightmargin=3pt,
    aboveskip=3pt,
    belowskip=3pt
}

\title{%
\parbox{\textwidth}{%
\centering
\rule{\linewidth}{2pt}\par\vspace{0.38cm}
Evo-B\&B: Behavior-Aware LLM-Guided Joint
Evolution of Branching and Node Selection
Policies for MILP
\par\vspace{-0.1cm}
\rule{\linewidth}{0.8pt}%
}%
}

\author{%
\normalfont
\makebox[\dimexpr\textwidth-2\tabcolsep\relax][c]{%
\begin{tabular}{@{}c@{\hspace{0.8em}}c@{\hspace{0.8em}}c@{}}
\shortstack[c]{
\textbf{Jiayu Chen}$^{1}$\\
{\footnotesize\rmfamily chenjiqyu@stu.xjtu.edu.cn}} &
\shortstack[c]{
\textbf{Chentong Chen}$^{1}$\\
{\footnotesize\rmfamily chengtong.chen@stu.xjtu.edu.cn}} &
\shortstack[c]{
\textbf{Yuhan Liu}$^{1}$\\
{\footnotesize\rmfamily LiuYuh@stu.xjtu.edu.cn}} \\[0.8em]
\shortstack[c]{
\textbf{Ye Fan}$^{2}$\\
{\footnotesize\rmfamily fanye@nwpu.edu.cn}} &
\shortstack[c]{
\textbf{Jianyong Sun}$^{1}$\\
{\footnotesize\rmfamily jy.sun@xjtu.edu.cn}} &
\shortstack[c]{
\textbf{Jialong Shi}$^{1,*}$\\
{\footnotesize\rmfamily jialong.shi@xjtu.edu.cn}} \\[0.8em]
\multicolumn{3}{c}{
\footnotesize
$^{1}$School of Mathematics and Statistics,
Xi'an Jiaotong University, Xi'an, China
}\\[0.12em]
\multicolumn{3}{c}{
\footnotesize
$^{2}$School of Electronics and Information,
Northwestern Polytechnical University, Xi'an, China
}
\end{tabular}%
}%
}

\iclrfinalcopy
\begin{document}

\maketitle
\pagestyle{plain}
\thispagestyle{plain}
\begingroup
\renewcommand{\thefootnote}{\fnsymbol{footnote}}
\footnotetext[1]{Corresponding author.}
\endgroup

\begin{abstract}
In branch-and-bound for Mixed-Integer Linear Programming (MILP), branching and node selection are two critical and tightly coupled decisions, yet most existing learning-based approaches focus on optimizing them separately. We propose Evo-B\&B, an LLM-guided evolutionary framework that jointly evolves executable branching and node selection policies using end-to-end SCIP solving feedback. To improve search efficiency and reduce redundant evaluations of behaviorally similar policies, Evo-B\&B introduces a behavior-aware multi-island evolutionary mechanism. It measures policy similarity from branching and node-selection behaviors on a fixed set of B\&B decision states and uses this similarity to preserve population diversity. We evaluate Evo-B\&B on four standard MILP benchmarks across multiple instance scales. Experimental results show that Evo-B\&B achieves competitive solving performance and generalizes well to larger-scale instances, while ablation studies further demonstrate the effectiveness of joint evolution and behavior-aware diversity management.
\end{abstract}
\section{Introduction}
\label{sec:introduction}
Mixed-Integer Linear Programming (MILP) is an important modeling paradigm in combinatorial optimization and operations research, with broad applications to real-world decision-making problems~\citep{bengio2021machine,gleixner2021miplib}. Since solving general MILPs is NP-hard, exact optimization of large-scale instances remains computationally challenging. Branch-and-Bound (B\&B) is the core framework underlying modern exact MILP solvers~\citep{achterberg2009scip,lodi2017learning}, where branching and node selection are two key search decisions. Branching determines how the current search space is partitioned, while node selection determines which open node is explored next~\citep{achterberg2005branching,he2014learning}. These two decisions continuously interact throughout the search and jointly affect the structure of the search tree and the overall solving efficiency~\citep{du2025learning}.

A variety of manually designed and data-driven methods have been developed for these search decisions. Traditional MILP solvers mainly rely on branching rules such as strong branching, pseudocost branching, and reliability branching, together with node-selection strategies such as best-bound and depth-first search~\citep{achterberg2005branching,achterberg2009scip}. More recently, machine learning has been widely applied to automate the design of branching and node-selection policies, including branching methods based on graph neural networks and search-tree information~\citep{gasse2019exact,gupta2020hybrid,zarpellon2021parameterizing,scavuzzo2022learning}, as well as learning-based node-selection methods~\citep{he2014learning,yilmaz2021study,zhang2025learning}. Recent work has further begun to consider branching and node selection jointly within a unified framework~\citep{du2025learning}.

Meanwhile, the program-generation capability of Large Language Models (LLMs) has opened a new direction for automated algorithm design. FunSearch~\citep{romeraparedes2024mathematical}, Evolution of Heuristics (EoH)~\citep{liu2024evolution}, and ReEvo~\citep{ye2024reevo} have shown that LLMs can be combined with program search and external evaluation feedback to iteratively generate and improve executable algorithms. This paradigm has also begun to extend to MILP solver design, including primal heuristics~\citep{zhang2025dhevo}, acceleration cuts~\citep{yazdani2025evocut}, branching policies~\mbox{\citep{hou2026llm4branch}}, and the programmatic evolution of multiple solver components~\citep{nie2026milpevo}. These studies demonstrate the potential of LLM-guided program evolution to directly search for explicit and executable solver logic based on actual solving performance.

Existing LLM-based policy evolution methods for MILP primarily focus on optimizing individual solver components, while the joint evolution of branching and node-selection policies remains underexplored. However, jointly searching over these two policies substantially enlarges the search space and can introduce considerable redundancy. Structural differences between programs do not necessarily translate into different decision behaviors~\citep{zhang2026rethinking}, and structurally distinct policies may still produce highly similar branching and node-selection decisions. Since each candidate policy must be evaluated through expensive end-to-end MILP solving, repeatedly exploring behaviorally similar policies can consume a significant portion of the limited search budget. Therefore, maintaining behaviorally meaningful diversity during joint evolution is a key challenge for improving search efficiency.

To address these challenges, we propose \textbf{Evo-B\&B}, an LLM-guided framework for jointly evolving branching and node-selection policies. Evo-B\&B represents the two decisions as a unified executable policy and jointly evaluates and optimizes them using end-to-end SCIP solving performance. To reduce behavioral redundancy, we introduce a behavior-aware multi-island evolutionary mechanism that measures policy similarity based on branching and node-selection behaviors under fixed B\&B decision states, thereby maintaining population diversity. In addition, Evo-B\&B combines LLM-driven structural search with selective parameter optimization, tuning only high-potential policies. Our main contributions are summarized as follows:
\begin{itemize}
\item We propose \textbf{Evo-B\&B}, which represents MILP branching and node selection as a unified executable policy and jointly evolves the two decisions using end-to-end SCIP solving feedback.
\item We introduce a behavior-aware multi-island evolutionary mechanism that uses the similarity of branching and node-selection behaviors to reduce redundant search, while selectively optimizing the parameters of high-potential policies.
\item We evaluate Evo-B\&B on four MILP benchmarks, including Setcover, Cauctions, Facilities, and Indset. The results demonstrate competitive solving performance and good generalization across different problem scales, while ablation studies further validate the effectiveness of behavior-aware diversity management and joint evolution.
\end{itemize}
\section{Related Work}
\label{sec:related_work}
Machine learning has been widely studied for improving branching decisions in B\&B. Early work approximated strong branching through supervised learning~\citep{khalil2016learning,alvarez2017machine}. Subsequent methods, including \citet{gasse2019exact}, \citet{gupta2020hybrid}, and TreeGate~\citep{zarpellon2021parameterizing}, improved branching policies from the perspectives of graph structure, efficient model design, and search-tree state information, respectively. Beyond neural approaches, Symb4CO~\citep{kuang2024symb4co} and GS4CO~\citep{kuang2024gs4co} further explored explicit and interpretable symbolic branching policies. Node selection has also been formulated as a learning problem~\citep{he2014learning,yilmaz2021study,zhang2025learning}, and recent work has begun to jointly consider branching and node selection~\citep{du2025learning}. In contrast, Evo-B\&B directly searches for both decision logics in a unified executable program space and evaluates them jointly using end-to-end MILP solving performance.

LLM-guided program evolution has recently emerged as a promising paradigm for automated algorithm design~\citep{romeraparedes2024mathematical,liu2024evolution,ye2024reevo}, and has gradually been extended to MILP solver design, including primal heuristics~\citep{zhang2025dhevo}, acceleration cuts~\citep{yazdani2025evocut}, branching policies~\citep{hou2026llm4branch}, and the program evolution of multiple solver components~\citep{nie2026milpevo}. Evo-B\&B jointly evolves branching and node selection, further incorporating behavior-aware multi-island population management based on policy similarity, together with selective parameter optimization to improve search efficiency.

\section{Method}
\label{method}

\begin{figure}[htb]
    \centering
    \includegraphics[width=\linewidth]{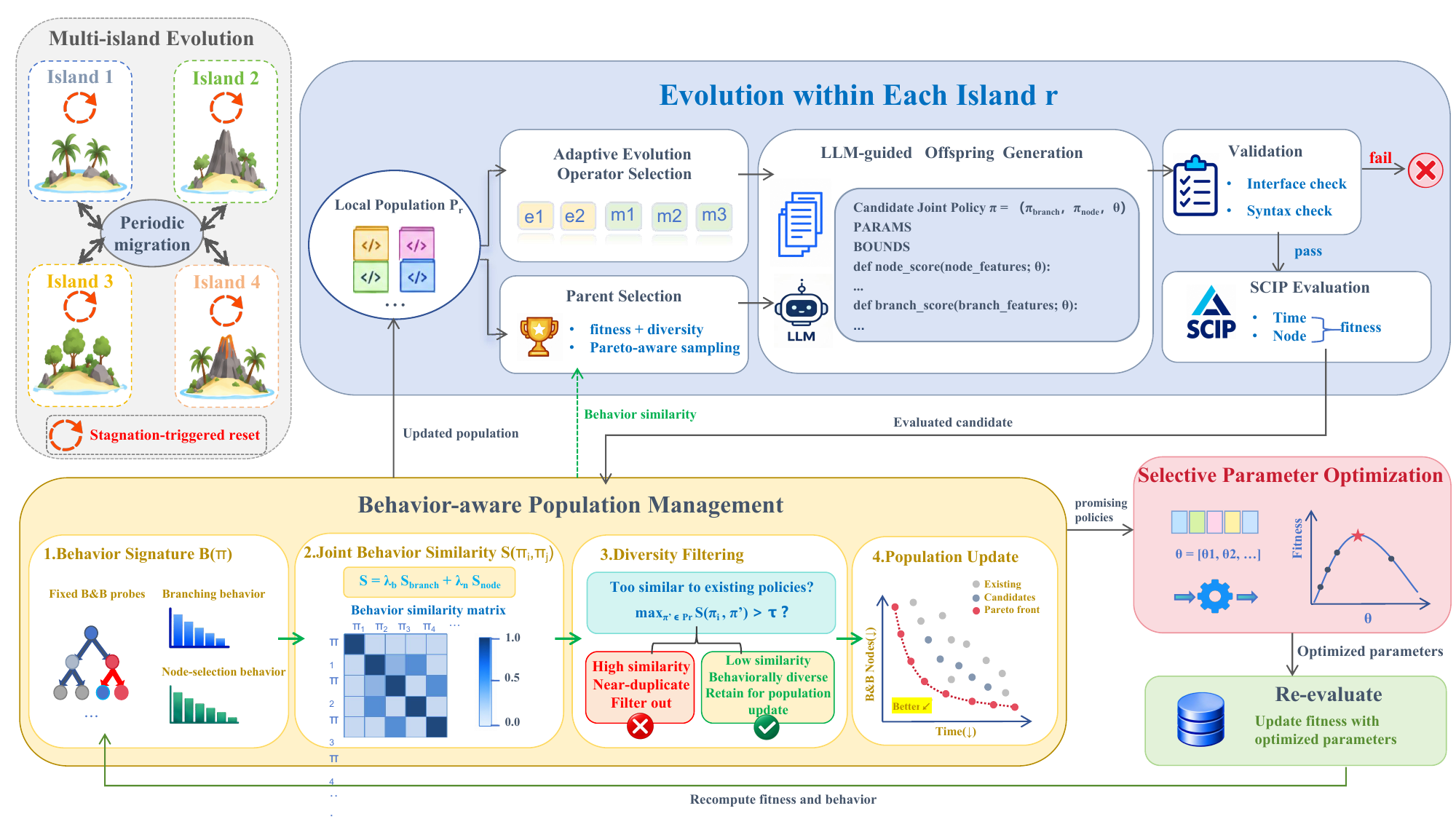}
    \caption{Overview of the Evo-B\&B framework. Evo-B\&B jointly evolves branching and node-selection policies through LLM-guided multi-island evolution, behavior-aware population management, and selective parameter optimization.}
    \label{fig:framework}
\end{figure}

We propose Evo-B\&B, an LLM-guided evolutionary framework for jointly optimizing branching-variable selection and node selection in MILP branch-and-bound. As illustrated in Figure~\ref{fig:framework}, Evo-B\&B represents the two decisions as a unified executable joint policy and improves it through behavior-aware multi-island evolution and selective parameter optimization. The framework integrates LLM-guided policy generation, behavior-aware diversity management, cross-island information exchange, and targeted parameter refinement into a unified evolutionary search process.

\subsection{Joint Policy Representation}

Evo-B\&B represents a branching policy and a node selection policy as a joint policy individual $\pi=(S_{\mathrm{branch}},S_{\mathrm{node}},\theta)$, where $S_{\mathrm{branch}}$ denotes the branching-variable scoring function, $S_{\mathrm{node}}$ denotes the node-selection scoring function, and $\theta$ denotes the continuous parameters of the policy.

During B\&B search, $S_{\mathrm{branch}}$ determines which candidate variable is selected for branching, thereby influencing how the current node is partitioned into child subproblems, while $S_{\mathrm{node}}$ determines the order in which open nodes are explored. Evo-B\&B separates structural search from numerical parameter optimization. The LLM generates the scoring logic of $S_{\mathrm{branch}}$ and $S_{\mathrm{node}}$, while the parameters $\theta$ are further optimized in the subsequent parameter tuning stage. Each LLM-generated joint policy is implemented as executable Python code and embedded into SCIP through a fixed interface~\citep{maher2016pyscipopt}.

\subsection{Behavior-aware Multi-Island Evolution}

\subsubsection{Behavior-aware Diversity Management}

During the policy search process of Evo-B\&B, candidate policies with different program structures may still induce similar B\&B decisions. Therefore, diversity management based solely on code structure may fail to identify behaviorally redundant policies~\citep{zhang2026rethinking}.
To address this issue, Evo-B\&B records how different policies score and select candidate branching variables and open nodes on a fixed set of B\&B decision states, and uses these records to measure behavioral similarity. The resulting similarity is further used for parent selection and population management to reduce behavioral redundancy and maintain population diversity.

\paragraph{Behavior Signature.}
Evo-B\&B constructs a behavior signature for each candidate joint policy to describe its preference under local B\&B decision states. Evo-B\&B first solves fixed instances with the default SCIP solver and collects a set of branching and node-selection decision probes during the solving process. Then, for each candidate policy $\pi$, Evo-B\&B computes its scores and selected candidates on these fixed probes. These scores are used only for behavior recording and do not affect the default SCIP solving process. Formally, the behavior signature is defined as
\begin{equation}
\mathcal{B}(\pi)=
\left(
\mathcal{B}_{\mathrm{branch}}(\pi),
\mathcal{B}_{\mathrm{node}}(\pi)
\right)
\label{eq:behavior_signature}
\end{equation}
where $\mathcal{B}_{\mathrm{branch}}(\pi)$ and $\mathcal{B}_{\mathrm{node}}(\pi)$ denote the sets of branching-decision records and node-selection records, respectively. Each behavior record contains not only the final selected candidate, but also the policy's relative preference over the entire candidate set.

\paragraph{Behavior Similarity.}
Given two policies $\pi_i$ and $\pi_j$, Evo-B\&B computes their behavior similarity by comparing their records on the same decision probes. The top-choice consistency $S_{\mathrm{top}}^q(\pi_i,\pi_j)$ measures whether the two policies select the same candidate on probe $q$, while the distribution similarity $S_{\mathrm{dist}}^q(\pi_i,\pi_j)$ measures whether their relative preferences over the whole candidate set are close. The behavior similarity on a single probe is defined as
\begin{equation}
S_q(\pi_i,\pi_j)
=
0.5 S_{\mathrm{top}}^q(\pi_i,\pi_j)
+
0.5 S_{\mathrm{dist}}^q(\pi_i,\pi_j)
\label{eq:probe_behavior_similarity}
\end{equation}
Evo-B\&B then aggregates the similarities over matched branching probes and node-selection probes, obtaining $S_{\mathrm{branch}}(\pi_i,\pi_j)$ and $S_{\mathrm{node}}(\pi_i,\pi_j)$. The final joint behavior similarity is defined as
\begin{equation}
S_{\mathrm{joint}}(\pi_i,\pi_j)
=
0.5 S_{\mathrm{branch}}(\pi_i,\pi_j)
+
0.5 S_{\mathrm{node}}(\pi_i,\pi_j)
\label{eq:joint_behavior_similarity}
\end{equation}
Details of decision probe construction, score normalization, and distribution similarity computation are provided in Appendix~\ref{app:behavior_similarity}.

\paragraph{Population Management.}
\label{sec:behavior_population}
Each island maintains a finite-capacity policy population $\mathcal{P}_r$. After a candidate policy $\pi_{\mathrm{new}}$ passes the safety validation and completes the end-to-end solving evaluation, Evo-B\&B computes its maximum behavioral similarity to the current population:
\begin{equation}
S_{\max}(\pi_{\mathrm{new}},\mathcal{P}_r)
=
\max_{\pi \in \mathcal{P}_r}
S_{\mathrm{joint}}(\pi_{\mathrm{new}},\pi)
\label{eq:max_behavior_similarity}
\end{equation}
When $S_{\max}>\tau$, $\pi_{\mathrm{new}}$ is regarded as behaviorally similar to an existing policy and is compared with its most similar policy using Pareto dominance. The comparison is based on the geometric-mean solving-time ratio and node-count ratio relative to default SCIP. If $\pi_{\mathrm{new}}$ is no worse in both objectives and strictly better in at least one, it replaces the most similar policy; otherwise, it is discarded as behaviorally redundant. 

When the population exceeds its capacity, Evo-B\&B performs non-dominated sorting using the same two objectives and retains policies sequentially according to their Pareto fronts. If a Pareto front cannot be fully retained within the remaining capacity, only the policies in that front are further selected by jointly considering policy quality and behavioral diversity. Given the set of already selected policies $\mathcal{S}$, the behavioral diversity of a candidate policy $\pi$ is defined as
\begin{equation}
D(\pi,\mathcal{S})
=
1-\max_{\pi'\in\mathcal{S}}
S_{\mathrm{joint}}(\pi,\pi')
\label{eq:behavior_diversity}
\end{equation}
The corresponding quality--diversity utility is defined as
\begin{equation}
U(\pi)
=
0.5\widehat{F}(\pi)
+
0.5D(\pi,\mathcal{S})
\label{eq:quality_diversity_utility}
\end{equation}
where $\widehat{F}(\pi)$ denotes the min--max normalized fitness within the current candidate set, and the definition of $F(\pi)$ is provided in Appendix~\ref{app:fitness}. When $\mathcal{S}=\emptyset$, we set $D(\pi,\mathcal{S})=0$, so the first policy is selected solely according to fitness.

\subsubsection{LLM-guided Multi-Island Evolution}

Evo-B\&B adopts a multi-island evolutionary framework, where policy search is distributed across multiple relatively independent local populations~\citep{chen2026tide}. In each evolutionary round, every island independently selects an evolution operator and constructs a parent pool. The LLM then generates a new joint policy according to the selected operator. After validity checking and end-to-end solving evaluation, the candidate policy is considered for insertion according to the behavior-aware population management mechanism described in Section~\ref{sec:behavior_population}.

\paragraph{Island Initialization.}
Let there be $K$ islands, and let $\mathcal{P}_r$ denote the local policy population maintained by the $r$-th island. Evo-B\&B initializes the islands with a set of diverse seed policies that satisfy the same joint policy interface. Each seed policy contains both a branching-variable scoring function and a node-selection scoring function, while differing in feature usage, scoring structure, and search preference.

\paragraph{Adaptive Operator Selection.}
\label{sec:operator_selection}
Evo-B\&B maintains an independent adaptive operator scheduler for each island, with the operator set
$\mathcal{A}=\{e_1,e_2,m_1,m_2,m_3\}$,
where $e_1$ and $e_2$ are crossover operators, and $m_1$, $m_2$, and $m_3$ are mutation operators. Following the operator-based evolutionary prompting paradigm of EoH~\citep{liu2024evolution}, each operator is associated with a dedicated prompt template that defines the corresponding policy transformation. The detailed prompt templates are provided in Appendix~\ref{app:prompt_design}.

At the beginning of evolution, previously unused operators are prioritized until each operator has received at least one round of feedback. Evo-B\&B then adopts a UCB-based adaptive scheduling rule~\citep{auer2002finite,dacosta2008adaptive,candan2013nonstationary}. If an operator has not been used during the most recent 20 candidate generations on the current island, it is prioritized for selection; otherwise, the operator is selected according to
\begin{equation}
a_t
=
\arg\max_{a\in\mathcal{A}}
\left(
\bar{R}_a
+
c\sqrt{\frac{2\log T}{N_a}}
\right)
\label{eq:operator_ucb}
\end{equation}
where $N_a$ denotes the number of times operator $a$ has been selected, $T$ denotes the cumulative number of operator selections on the current island, $c$ is the exploration coefficient, and $\bar{R}_a$ is the historical average reward of operator $a$.

The reward of an operator is defined as the fitness improvement of the generated offspring over its primary parent:
\begin{equation}
R_a
=
F(\pi_{\mathrm{new}})
-
F(\pi_{\mathrm{par}})
\label{eq:operator_reward}
\end{equation}
where $F(\cdot)$ denotes the end-to-end fitness defined in Appendix~\ref{app:fitness}. If the generated policy fails the validity or safety checks, or fails during solver evaluation, the corresponding operator receives a negative reward.

\paragraph{Parent Selection.}

After determining the evolution operator, Evo-B\&B constructs a parent pool from the current island population. The policy with the highest fitness is first added to the pool, while the remaining candidates are ranked by non-dominated sorting according to their solving-time ratio and node-count ratio relative to default SCIP and selected in Pareto-front order. If a front cannot be fully retained within the remaining capacity, a quality--diversity criterion is further applied. For a candidate policy $\pi$, the parent-selection utility is defined as
\begin{equation}
U_p(\pi)
=
0.5\widehat{F}(\pi)
+
0.5
\left(
1-\max_{\pi'\in\mathcal{R}_r}
S_{\mathrm{joint}}(\pi,\pi')
\right)
\label{eq:parent_selection_utility}
\end{equation}
where $\widehat{F}(\pi)$ denotes the normalized end-to-end fitness, and $\mathcal{R}_r$ denotes the set of already selected parents. If the current island does not contain enough valid policies, additional parents are selected from other islands using the same rule. After the parent pool is constructed, one policy is randomly chosen as the primary parent. Mutation operators modify only the primary parent, whereas crossover operators use the primary parent together with up to two auxiliary parents to generate a new policy.

\paragraph{LLM-guided Offspring Generation and Evaluation.}

Given the selected evolution operator and parent policies, Evo-B\&B constructs a prompt containing the parent code, design rationale, solving feedback, and operator-specific instructions. The LLM then generates a complete joint policy that includes both a branching-variable scoring function and a node-selection scoring function. Detailed operator prompt templates are provided in Appendix~\ref{app:prompt_design}.

The generated candidate policy first undergoes interface and safety checks before being integrated into SCIP for end-to-end evaluation on the training set, with its fitness computed according to Appendix~\ref{app:fitness}. Evo-B\&B subsequently constructs its behavior signature and updates the current island population using the behavior-aware population management mechanism described in Section~\ref{sec:behavior_population}. Finally, the evaluation results are used to update the reward statistics of the corresponding evolution operator.

\subsubsection{Island Migration and Reset}

\paragraph{Island Migration.}

Evo-B\&B periodically performs information migration across islands. Rather than directly copying policy individuals, the migration process extracts design information from high-quality policies that are behaviorally different from the target population and stores it in the migration memory of the target island for subsequent policy generation and reset.

When migration is triggered, each island is considered as the target island in turn. For a target population $\mathcal{P}_r$, let $\mathcal{M}_r$ denote the set of candidate policies from the other islands. For each $\pi\in\mathcal{M}_r$, Evo-B\&B jointly considers its fitness and behavioral difference from the target population, and defines the migration utility as
\begin{equation}
U_{\mathrm{mig}}^r(\pi)
=
\widehat{F}_r(\pi)
+
\left(
1-
\max_{\pi'\in\mathcal{P}_r}
S_{\mathrm{joint}}(\pi,\pi')
\right)
\label{eq:migration_utility}
\end{equation}
where $\widehat{F}_r(\pi)$ denotes the min--max normalized fitness of policy $\pi$ within the candidate set $\mathcal{M}_r$. Evo-B\&B ranks candidate policies according to $U_{\mathrm{mig}}^r(\pi)$ and stores the design information of the top-ranked policies in the migration memory of the target island as auxiliary context for subsequent LLM-guided policy generation.

\paragraph{Island Reset.}

When an island shows no performance improvement for $H$ consecutive evolutionary rounds, Evo-B\&B triggers an island reset. During the reset, the best policy of the island is retained, the globally best policy is introduced, and two policies from other islands with both strong solving performance and substantial behavioral difference are selected as external migrants. Unlike periodic information migration, the reset process directly transfers policy individuals.

After the reset, the stagnation counter is cleared, and the fusion-reset operator $r_1$ is used for the next candidate generation. The corresponding reset prompt includes the complete code of the locally best and globally best policies, together with summary information of the external migrants, including their source islands, fitness values, solving-time ratios, and node-count ratios. This information guides the LLM to integrate search experience from different islands and generate a new joint policy.

\subsection{Selective Parameter Optimization}
LLM-based evolution primarily searches the program structure of the joint policy, while its continuous parameters can also affect solving performance. Optimizing every generated policy would require substantial additional MILP evaluations. Evo-B\&B therefore performs local parameter optimization only for high-potential policies~\citep{chen2026tide}. For a newly generated policy $\pi$ on island $r$, tuning is triggered only if it passes validity checks and end-to-end evaluation, is retained in the island population, and achieves a pre-tuning fitness of at least $95\%$ of the current island-best fitness.

During tuning, the code structures of $S_{\mathrm{branch}}$ and $S_{\mathrm{node}}$ remain fixed, and only the continuous parameter vector $\theta$ is optimized. Let $[l_j,u_j]$ denote the feasible range of parameter $\theta_j$. Candidate parameters are generated by bounded Gaussian mutation~\citep{beyer2002evolution}:
\begin{equation}
\theta'_j
=
\operatorname{clip}
\left(
\theta_j + z_j\epsilon_j,\,
l_j,\,
u_j
\right),
\qquad
z_j\sim\mathrm{Bernoulli}(p),\quad
\epsilon_j\sim
\mathcal{N}\left(0,\sigma_t^2(u_j-l_j)^2\right)
\label{eq:parameter_perturbation}
\end{equation}
where $p=0.5$ is the probability of perturbing each parameter. Each tuning procedure performs $T_{\mathrm{tune}}=10$ local search steps. With the base scale set to $\sigma_0=0.08$, the perturbation scale decreases over the search as
\begin{equation}
\sigma_t
=
\sigma_0
\left(
1-0.7\frac{t}{T_{\mathrm{tune}}}
\right),
\qquad
t=1,\ldots,T_{\mathrm{tune}}
\label{eq:parameter_sigma_decay}
\end{equation}
At each tuning step, a new parameter vector is generated from the current best one and evaluated on the same training instances. The candidate replaces the current best parameters only if it achieves a higher fitness. After $T_{\mathrm{tune}}$ steps, the best parameter vector found during tuning is returned.

\section{Experiments}
\label{sec:experiments}

\subsection{Experimental Settings}
\label{sec:experimental_settings}
\paragraph{Benchmarks.}
We evaluate Evo-B\&B on four classical MILP benchmark problems: Set Covering~\citep{balas1980set}, Combinatorial Auctions~\citep{leyton2000multi}, Capacitated Facility Location~\citep{cornuejols1991capacitated}, and Maximum Independent Set~\citep{bergman2016decision}. For convenience, we refer to these four benchmark classes as Setcover, Cauctions, Facilities, and Indset, respectively.
Following prior MILP solving studies~\citep{gasse2019exact,gupta2020hybrid}, we construct instances at three difficulty levels for each benchmark: Easy, Medium, and Hard. Easy instances are used for policy evolution, while Medium and Hard instances are used to evaluate the generalization of the discovered policies to larger-scale instances. Detailed benchmark configurations and parameter settings are provided in Appendix~\ref{app:benchmark_details}.

\paragraph{Baselines.}
\label{sec:baselines}
We compare Evo-B\&B with several representative methods for improving B\&B search, including traditional solver heuristics, learning-based methods, and LLM-based methods. Traditional baselines include the default SCIP strategy (Default), Reliability Pseudocost Branching (RPB), and Full Strong Branching (FSB)~\citep{achterberg2005branching,achterberg2009scip}. Learning-based methods include GNN~\citep{gasse2019exact}, Hybrid~\citep{gupta2020hybrid}, TreeGate~\citep{zarpellon2021parameterizing}, and Symb4CO~\citep{kuang2024symb4co}, where Symb4CO searches for explicit and interpretable symbolic branching policies. LLM-based methods include FunSearch~\citep{romeraparedes2024mathematical} and LLM4Branch~\citep{hou2026llm4branch}. Since the original FunSearch is not designed for joint branching and node-selection policy search in MILP, we adapt it to the same joint-policy setting as Evo-B\&B and evaluate the generated policies end-to-end through the same solver interface. LLM4Branch uses LLMs to generate and iteratively improve executable branching policies.

\paragraph{Training Settings.}
We independently evolve one policy for each benchmark using $10$ Easy instances as the training set, with a solving time limit of $60$ seconds per instance. The evolutionary search is driven by Qwen3-Max. We use $4$ islands, each with a maximum population size of $8$, and generate one offspring per round. The search terminates when either $150$ rounds or a $10$-hour time budget is reached, whichever comes first.
The behavioral-similarity threshold is set to $\tau=0.98$, and cross-island migration is triggered every $10$ rounds. An island reset is triggered if its best fitness does not improve for $8$ consecutive rounds. For eligible high-potential policies, we further perform $10$ steps of local parameter optimization, with a base perturbation scale of $\sigma_0=0.08$ and a parameter perturbation probability of $0.5$. Additional details on the LLM configuration, behavior probe construction, and evolutionary hyperparameters are provided in Appendix~\ref{app:training_details}.

\paragraph{Evaluation Settings.}
Evo-B\&B is implemented with SCIP 10.0.1 and PySCIPOpt 6.1.0. For existing baseline methods, we follow their original implementations. All methods are evaluated on the same test instances under the same per-instance time limits. Specifically, the Easy, Medium, and Hard settings each contain $80$ test instances, with time limits of $60$, $600$, and $3600$ seconds per instance, respectively. We report three metrics. \textbf{Time} and \textbf{Nodes} denote the arithmetic mean of solving time and the number of B\&B nodes over all test instances, respectively, with lower values being better. \textbf{Wins} denotes the number of instances on which a method achieves the fastest solving time among all compared methods; for instances with tied best performance, each tied method is counted as one win. Additional parameter settings are provided in Appendix~\ref{app:training_details}.
\subsection{Main Results}
\label{sec:main_results}
The main results of Evo-B\&B on the four MILP benchmark classes are reported in Table~\ref{tab:main_results}. Across the $12$ problem--scale settings, Evo-B\&B achieves the lowest average solving time in $6$ settings and 
\clearpage

\begin{table}[H]
\centering
\caption{
Main results on four MILP benchmarks.
Lower Time and Nodes are better, while higher Wins is better.
Best results are shown in bold.
}
\label{tab:main_results}

\vspace{-0.25em}

\small
\setlength{\tabcolsep}{4.2pt}
\renewcommand{\arraystretch}{1.0}

\begin{tabular*}{\linewidth}{@{\extracolsep{\fill}}lrrrrrrrrr@{}}
\toprule
Setcover:
& \multicolumn{3}{c}{Easy}
& \multicolumn{3}{c}{Medium}
& \multicolumn{3}{c}{Hard} \\
\cmidrule(lr){1-4}
\cmidrule(lr){5-7}
\cmidrule(lr){8-10}

Method
& Time(s)$\downarrow$ & Nodes$\downarrow$ & Wins$\uparrow$
& Time(s)$\downarrow$ & Nodes$\downarrow$ & Wins$\uparrow$
& Time(s)$\downarrow$ & Nodes$\downarrow$ & Wins$\uparrow$ \\
\midrule

FSB
& 31.52 & \textbf{75.80} & 0/63
& 457.31 & \textbf{389.31} & 0/34
& 3557.49 & \textbf{1610.46} & 0/2 \\

RPB
& 8.41 & 323.05 & 0/80
& 113.72 & 8982.70 & 0/78
& 1841.47 & 132637.61 & 0/54 \\

Default
& 8.77 & 323.05 & 0/80
& 115.51 & 8877.84 & 0/78
& 1839.25 & 131968.26 & 0/54 \\

\cmidrule(lr){1-4}
\cmidrule(lr){5-7}
\cmidrule(lr){8-10}

GNN
& 17.13 & 407.16 & 0/73
& 346.66 & 4318.96 & 0/44
& 3487.07 & 21551.94 & 0/7 \\

Hybrid
& 35.00 & 158.62 & 0/57
& 432.05 & 1213.90 & 0/39
& 3445.38 & 8280.55 & 0/10 \\

TreeGate
& 29.21 & 777.10 & 0/66
& 374.74 & 6961.51 & 0/46
& 3266.46 & 58041.30 & 0/19 \\

Symb4CO
& 13.12 & 415.94 & 0/79
& 173.74 & 6565.08 & 0/74
& 2343.18 & 77236.39 & 0/46 \\

\cmidrule(lr){1-4}
\cmidrule(lr){5-7}
\cmidrule(lr){8-10}

FunSearch
& 4.74 & 791.25 & 0/80
& 126.94 & 18557.96 & 0/75
& 2320.65 & 233085.12 & 0/41 \\

LLM4Branch
& 3.59 & 423.69 & 36/80
& \textbf{72.46} & 8174.41 & \textbf{42/79}
& \textbf{1582.26} & 137975.27 & \textbf{47/61} \\

Ours
& \textbf{3.44} & 499.48 & \textbf{44/80}
& 80.11 & 11801.01 & 37/78
& 1717.87 & 205433.60 & 14/56 \\

\bottomrule
\end{tabular*}

\par\vspace{0.55\baselineskip}

\begin{tabular*}{\linewidth}{@{\extracolsep{\fill}}lrrrrrrrrr@{}}
\toprule
Cauctions:
& \multicolumn{3}{c}{Easy}
& \multicolumn{3}{c}{Medium}
& \multicolumn{3}{c}{Hard} \\
\cmidrule(lr){1-4}
\cmidrule(lr){5-7}
\cmidrule(lr){8-10}

Method
& Time(s)$\downarrow$ & Nodes$\downarrow$ & Wins$\uparrow$
& Time(s)$\downarrow$ & Nodes$\downarrow$ & Wins$\uparrow$
& Time(s)$\downarrow$ & Nodes$\downarrow$ & Wins$\uparrow$ \\
\midrule

FSB
& 9.23 & 65.94 & 0/80
& 337.76 & \textbf{597.58} & 0/58
& 3414.62 & \textbf{2262.48} & 0/8 \\

RPB
& 2.18 & \textbf{32.49} & 0/80
& 25.79 & 2678.89 & 0/80
& 334.27 & 32773.91 & 0/80 \\

Default
& 2.20 & \textbf{32.49} & 0/80
& 24.87 & 2678.89 & 7/80
& 293.25 & 32773.91 & 16/80 \\

\cmidrule(lr){1-4}
\cmidrule(lr){5-7}
\cmidrule(lr){8-10}

GNN
& 5.16 & 943.50 & 0/79
& 279.11 & 27847.29 & 0/60
& 2928.13 & 264826.50 & 0/24 \\

Hybrid
& 21.25 & 102.31 & 0/72
& 381.31 & 1242.26 & 0/53
& 3175.21 & 9184.00 & 0/22 \\

TreeGate
& 27.49 & 398.84 & 0/65
& 316.75 & 5114.66 & 0/62
& 2199.87 & 44011.11 & 0/56 \\

Symb4CO
& 2.10 & 121.45 & 0/80
& 34.08 & 2652.74 & 0/80
& 707.16 & 42069.65 & 0/80 \\

\cmidrule(lr){1-4}
\cmidrule(lr){5-7}
\cmidrule(lr){8-10}

FunSearch
& 1.20 & 143.66 & 11/80
& 326.41 & 6173.21 & 0/38
& 3516.65 & 19657.83 & 0/2 \\

LLM4Branch
& \textbf{0.94} & 212.20 & \textbf{42/80}
& \textbf{23.32} & 6260.14 & 32/80
& \textbf{286.70} & 58427.75 & \textbf{63/80} \\

Ours
& 0.98 & 129.62 & 27/80
& 41.42 & 4137.64 & \textbf{41/78}
& 1011.31 & 59289.40 & 1/73 \\

\bottomrule
\end{tabular*}

\par\vspace{0.55\baselineskip}

\begin{tabular*}{\linewidth}{@{\extracolsep{\fill}}lrrrrrrrrr@{}}
\toprule
Facilities:
& \multicolumn{3}{c}{Easy}
& \multicolumn{3}{c}{Medium}
& \multicolumn{3}{c}{Hard} \\
\cmidrule(lr){1-4}
\cmidrule(lr){5-7}
\cmidrule(lr){8-10}

Method
& Time(s)$\downarrow$ & Nodes$\downarrow$ & Wins$\uparrow$
& Time(s)$\downarrow$ & Nodes$\downarrow$ & Wins$\uparrow$
& Time(s)$\downarrow$ & Nodes$\downarrow$ & Wins$\uparrow$ \\
\midrule

FSB
& 43.92 & \textbf{42.91} & 0/48
& 443.48 & \textbf{103.65} & 0/45
& 2583.80 & \textbf{149.99} & 0/44 \\

RPB
& 36.68 & 54.11 & 0/57
& 302.34 & 201.21 & 0/63
& 1191.48 & 245.94 & 0/78 \\

Default
& 36.61 & 55.44 & 0/57
& 304.91 & 197.45 & 0/63
& 1187.77 & 247.69 & 0/78 \\

\cmidrule(lr){1-4}
\cmidrule(lr){5-7}
\cmidrule(lr){8-10}

GNN
& 27.07 & 194.34 & 0/66
& 183.74 & 616.41 & 0/75
& 661.07 & 587.04 & 0/80 \\

Hybrid
& 29.33 & 161.04 & 0/64
& 135.97 & 588.67 & 7/78
& 420.18 & 508.31 & 23/80 \\

TreeGate
& 47.58 & 259.13 & 0/34
& 466.62 & 681.91 & 0/37
& 2373.98 & 869.58 & 0/56 \\

Symb4CO
& 30.51 & 163.59 & 0/66
& 218.90 & 389.75 & 0/71
& 966.10 & 366.49 & 0/79 \\

\cmidrule(lr){1-4}
\cmidrule(lr){5-7}
\cmidrule(lr){8-10}

FunSearch
& 25.55 & 288.49 & 0/67
& 189.76 & 744.96 & 0/76
& \textbf{411.59} & 712.51 & 16/80 \\

LLM4Branch
& 19.94 & 275.65 & 11/74
& 125.51 & 622.38 & 5/80
& 486.45 & 490.85 & 6/80 \\

Ours
& \textbf{17.05} & 378.29 & \textbf{65/74}
& \textbf{97.29} & 758.96 & \textbf{68/80}
& 418.92 & 720.86 & \textbf{35/80} \\

\bottomrule
\end{tabular*}

\par\vspace{0.55\baselineskip}

\begin{tabular*}{\linewidth}{@{\extracolsep{\fill}}lrrrrrrrrr@{}}
\toprule
Indset:
& \multicolumn{3}{c}{Easy}
& \multicolumn{3}{c}{Medium}
& \multicolumn{3}{c}{Hard} \\
\cmidrule(lr){1-4}
\cmidrule(lr){5-7}
\cmidrule(lr){8-10}

Method
& Time(s)$\downarrow$ & Nodes$\downarrow$ & Wins$\uparrow$
& Time(s)$\downarrow$ & Nodes$\downarrow$ & Wins$\uparrow$
& Time(s)$\downarrow$ & Nodes$\downarrow$ & Wins$\uparrow$ \\
\midrule

FSB
& 60.01 & \textbf{2.44} & 0/0
& 599.79 & \textbf{12.16} & 0/1
& 3600.06 & \textbf{24.39} & 0/0 \\

RPB
& 27.68 & 1122.59 & 0/74
& 181.32 & 8439.46 & 0/76
& 2487.65 & 80719.80 & 0/48 \\

Default
& 26.92 & 1133.20 & 0/74
& 174.39 & 8534.98 & 0/76
& 2435.89 & 84178.53 & 0/48 \\

\cmidrule(lr){1-4}
\cmidrule(lr){5-7}
\cmidrule(lr){8-10}

GNN
& 10.09 & 638.69 & 25/78
& 99.65 & 5658.26 & 8/77
& 1929.23 & 71590.84 & 6/54 \\

Hybrid
& 47.93 & 181.59 & 0/30
& 581.90 & 1366.00 & 0/3
& 3600.21 & 9517.02 & 0/0 \\

TreeGate
& 50.47 & 603.49 & 0/23
& 501.06 & 6480.60 & 0/24
& 3533.28 & 33161.96 & 0/5 \\

Symb4CO
& 36.47 & 468.31 & 0/69
& 190.24 & 4923.21 & 0/75
& 2515.07 & 47618.78 & 0/42 \\

\cmidrule(lr){1-4}
\cmidrule(lr){5-7}
\cmidrule(lr){8-10}
FunSearch
& 11.15 & 381.23 & 0/77
& 137.26 & 2385.10 & 0/69
& 2708.26 & 12368.84 & 0/23 \\

LLM4Branch
& 34.83 & 4862.73 & 2/50
& 469.51 & 53825.64 & 0/25
& 3614.31 & 195207.67 & 0/1 \\

Ours
& \textbf{7.71} & 434.40 & \textbf{52/79}
& \textbf{54.72} & 3040.60 & \textbf{72/80}
& \textbf{1458.65} & 55541.68 & \textbf{59/63} \\

\bottomrule
\end{tabular*}

\end{table}
the highest Wins in $8$ settings. More specifically, Evo-B\&B attains both the lowest solving time and the highest Wins on Setcover-Easy. On Facilities, it achieves the lowest solving time on the Easy and Medium scales and the highest Wins across all three scales. On Indset, it achieves both the lowest solving time and the highest Wins on Easy, Medium, and Hard.
In terms of node count, FSB explores fewer B\&B nodes in some settings, but a lower node count does not necessarily translate into a shorter solving time. Evo-B\&B also does not always achieve the lowest node count, yet it still obtains lower solving times and higher Wins in multiple settings. This indicates that node count alone does not fully reflect solving efficiency, whereas end-to-end solving time provides a more direct measure of policy performance in practice.
\subsection{Ablation Studies}
\label{sec:ablation_studies}
\begin{table}[t]
\centering
\caption{
Ablation results on Setcover and Indset.
Lower Time and Nodes are better, while higher Wins is better.
Best results are shown in bold.
}
\label{tab:ablation}

\small
\setlength{\tabcolsep}{5.0pt}
\renewcommand{\arraystretch}{1.0}

\begin{tabular*}{\linewidth}
{@{\extracolsep{\fill}}l|rrr|rrr@{}}
\toprule
Setcover:
& \multicolumn{3}{c|}{Easy}
& \multicolumn{3}{c}{Medium} \\

Variant
& Time(s)$\downarrow$ & Wins$\uparrow$ & Nodes$\downarrow$
& Time(s)$\downarrow$ & Wins$\uparrow$ & Nodes$\downarrow$ \\
\midrule

No similarity
& 5.15 & 14/80 & 544.5
& 114.10 & 5/77 & 12937.6 \\

AST
& 3.66 & 13/80 & 520.5
& 95.63 & 8/77 & 13390.2 \\

\midrule

Node only
& 10.37 & 0/78 & \textbf{299.4}
& 236.01 & 0/58 & \textbf{4069.4} \\

Branch only
& 7.98 & 0/79 & 494.2
& 152.91 & 0/73 & 7739.6 \\

Independent Node+Branch
& 9.50 & 0/78 & 471.2
& 242.90 & 0/58 & 4322.0 \\

\midrule

Ours
& \textbf{3.44} & \textbf{61/80} & 499.5
& \textbf{80.11} & \textbf{66/78} & 11801.0 \\

\bottomrule
\end{tabular*}

\vspace{0.6\baselineskip}

\begin{tabular*}{\linewidth}
{@{\extracolsep{\fill}}l|rrr|rrr@{}}
\toprule
Indset:
& \multicolumn{3}{c|}{Easy}
& \multicolumn{3}{c}{Medium} \\

Variant
& Time(s)$\downarrow$ & Wins$\uparrow$ & Nodes$\downarrow$
& Time(s)$\downarrow$ & Wins$\uparrow$ & Nodes$\downarrow$ \\
\midrule

No similarity
& 14.97 & 1/77 & 465.1
& 128.11 & 0/73 & \textbf{2896.1} \\

AST
& 7.86 & 26/79 & \textbf{420.8}
& 59.19 & 22/80 & 3205.3 \\

\midrule

Node only
& 34.02 & 0/63 & 975.6
& 270.31 & 0/69 & 7175.6 \\

Branch only
& 18.34 & 0/78 & 602.1
& 123.70 & 0/75 & 3883.5 \\

Independent Node+Branch
& 19.25 & 0/75 & 534.4
& 139.56 & 0/76 & 3626.2 \\

\midrule

Ours
& \textbf{7.71} & \textbf{53/79} & 434.4
& \textbf{54.72} & \textbf{59/80} & 3040.6 \\

\bottomrule
\end{tabular*}
\end{table}

We conduct ablation studies on Setcover and Indset to analyze the effects of behavior similarity and joint evolution of branching and node-selection policies, with the results reported in Table~\ref{tab:ablation}. The compared variants include removing the similarity mechanism (No similarity), replacing behavior similarity with AST-based code-structure similarity (AST), evolving only the node-selection policy (Node only), evolving only the branching policy (Branch only), and evolving the two policies independently before combining them for solving (Independent Node+Branch).
\paragraph{Effect of Behavior Similarity.}
Compared with No similarity and AST, the full Evo-B\&B achieves the lowest average solving time and the highest Wins in all four settings. These results show that, compared with using no similarity management or relying only on program-structure similarity, behavioral similarity defined over actual B\&B decisions is better suited for diversity management during evolutionary search and leads to improved end-to-end solving performance.
\paragraph{Effect of Joint Evolution.}
Compared with Node only, Branch only, and Independent Node+Branch, Evo-B\&B achieves the lowest average solving time and the highest Wins across all four experimental settings. In particular, Independent Node+Branch underperforms joint evolution in all settings, indicating that optimizing branching and node-selection policies separately and then combining them cannot fully exploit the coupling between the two decisions in B\&B search.
\section{Conclusion}

We propose Evo-B\&B, an LLM-guided joint evolutionary framework for MILP branch-and-bound solving that represents branching and node-selection policies as a unified executable policy and optimizes them using end-to-end SCIP solving feedback. To reduce behavioral redundancy during joint search, Evo-B\&B combines behavior-aware multi-island evolution, cross-island information exchange, and selective parameter optimization.
Experiments on four MILP benchmarks show that Evo-B\&B achieves strong solving performance across different problem scales, while ablation studies further validate the effectiveness of behavior-aware diversity management and joint evolution. Future work will extend the framework to additional solver components and explore more efficient policy evaluation and migration mechanisms to reduce evolutionary search cost and improve generalization.

\section*{AI Use Statement}

During the preparation of this paper, we used AI tools to assist with English language polishing, sentence revision, code implementation and debugging, and checks of LaTeX and citation formatting. AI tools were also used to provide feedback during the development and refinement of the research methodology and experimental design. They were not used to directly generate experimental results or to make final scientific decisions. All experimental results were obtained through our own implementations and experiments, and all analyses, claims, and conclusions were independently verified and determined by the authors. All AI-assisted text and code were manually reviewed, tested, and revised where necessary. We take full responsibility for the final content of this work.
\bibliography{iclr2027_conference}
\bibliographystyle{iclr2027_conference}

\appendix
\section{Preliminary}
\label{app:preliminary}

\subsection{Mixed-Integer Linear Programming and Branch-and-Bound}

A mixed-integer linear programming (MILP) problem can be formulated as
\[
\min_{\mathbf{x}} \mathbf{c}^{\top}\mathbf{x}
\quad \mathrm{s.t.}\quad
A\mathbf{x}\leq \mathbf{b},\quad x_j\in\mathbb{Z},\ j\in\mathcal{I}
\]

Here, $\mathbf{x}$ denotes the decision variables, $\mathbf{c}$ denotes the objective coefficients, $A$ and $\mathbf{b}$ denote the constraint matrix and the right-hand-side vector, respectively, and $\mathcal{I}$ denotes the index set of integer variables.

Modern MILP solvers typically use Branch-and-Bound (B\&B) as the core solving framework \citep{achterberg2009scip}. B\&B obtains bounds by solving the LP relaxation at each node and recursively partitions the search space by branching on fractional solutions. During the search process, the solver maintains a set of unexplored open nodes until all nodes have either been processed or pruned, thereby certifying the optimality of the incumbent solution.

\subsection{Branching and Node Selection}

During B\&B solving, the branching policy and the node selection policy are two key decision components that affect search efficiency. The branching policy selects a fractional variable from the LP relaxation at the current node for branching, thereby determining how the current node is partitioned into new subproblems. The node selection policy selects the next node to process from the set of open nodes maintained by the solver, thereby determining the traversal order of the search tree \citep{achterberg2005branching,linderoth1999search}.

These two decisions operate at different stages of the B\&B search process. The branching policy mainly affects how the search tree is generated, including the partitioning of child nodes and the structure of the subsequent search space. In contrast, the node selection policy mainly affects how the search tree is traversed by determining which open nodes are explored first, and consequently influences the evolution of the primal and dual bounds. Therefore, the two decisions jointly affect the structure of the B\&B search tree, the search order, and the overall solving efficiency.

Although branching and node selection are typically designed as two relatively independent components, they are closely coupled during the search process. The branching policy determines which child nodes are generated, while the open-node set faced by the node selection policy is progressively constructed by previous branching decisions. Meanwhile, the node visitation order also changes the search states in which subsequent branching decisions are made. Therefore, optimizing only one of these decision components may fail to fully exploit their interaction, whereas jointly optimizing branching and node selection policies has the potential to further improve the overall MILP solving performance.

\section{Behavior Probe Construction and Similarity Calculation}
\label{app:behavior_similarity}

\subsection{Behavior Probe Construction}
\label{app:behavior_probe_construction}

Evo-B\&B uses decision probes to characterize the local decision behavior of candidate policies during B\&B search. We construct behavior probes using instances with seeds $0$--$2$ from both the Easy and Medium scales, resulting in $3$ fixed instances per scale and $6$ probe instances in total. For each instance, Default SCIP is used to generate a fixed reference search trajectory, from which at most $50$ branching states and $50$ node-selection states are recorded. All candidate policies are then evaluated on these fixed decision states. Their scores and the choices induced by these scores are used only to construct behavior records and do not alter the reference solving process of Default SCIP. This allows different policies to be compared under the same B\&B decision states, avoiding the influence of different search trajectories on behavior similarity.

Decision probes are divided into branching probes and node-selection probes. For the $t$-th branching probe $q_t^b$, let the set of candidate branching variables in the current LP relaxation be
$C_t=\{v_1,\ldots,v_m\}$.
The probe is identified by the instance, decision type, position in the reference solving process, and the number of candidate variables $m$. Given a candidate policy $\pi$, Evo-B\&B scores all candidate variables in this state and obtains
$\mathbf{s}_{q_t^b}(\pi)\in\mathbb{R}^{m}$.
The branching candidate induced by these scores is
\begin{equation}
a_{q_t^b}(\pi)
=
\arg\max_i \mathbf{s}_{q_t^b,i}(\pi)
\end{equation}
The behavior record of policy $\pi$ on this branching probe is defined as
\begin{equation}
r_{q_t^b}(\pi)
=
\left(
\mathbf{s}_{q_t^b}(\pi),
a_{q_t^b}(\pi)
\right)
\label{eq:branch_behavior_record}
\end{equation}

For the $t$-th node-selection probe $q_t^n$, let the current set of open nodes be
$N_t=\{n_1,\ldots,n_k\}$.
Similarly, the probe is identified by the instance, decision type, position in the reference solving process, and the number of open nodes $k$. The candidate policy $\pi$ scores all open nodes in this state, obtaining
$\mathbf{s}_{q_t^n}(\pi)\in\mathbb{R}^{k}$.
The node induced by these scores is
\begin{equation}
a_{q_t^n}(\pi)
=
\arg\max_i \mathbf{s}_{q_t^n,i}(\pi)
\end{equation}
The behavior record of policy $\pi$ on this node-selection probe is defined as
\begin{equation}
r_{q_t^n}(\pi)
=
\left(
\mathbf{s}_{q_t^n}(\pi),
a_{q_t^n}(\pi)
\right)
\label{eq:node_behavior_record}
\end{equation}

Two behavior records are treated as matched decision probes only when they correspond to the same instance, decision type, position in the reference solving process, and candidate-set size. Since all candidate policies are evaluated on the same reference states generated by Default SCIP, valid policies generally share a consistent probe sequence. For the small number of probes missing due to scoring failures or incomplete behavior records, Evo-B\&B computes behavior similarity only over the matched probes shared by both policies.

The behavior duplicate filter is enabled throughout the evolutionary process, with the similarity threshold set to
$\tau_{\mathrm{dup}}=0.98$.
When the maximum behavioral similarity between a new candidate and an existing policy in the current island exceeds this threshold, their solving-time and node-count performance on the training set are further compared to determine whether the new candidate should be retained, preventing behaviorally redundant policies from occupying the limited population capacity.

\subsection{Probe-level Similarity}
\label{app:probe_level_similarity}

Given two policies $\pi_i$ and $\pi_j$, Evo-B\&B first compares their local decision behavior on a decision probe $q$. Let their score vectors on this probe be $\mathbf{s}_q(\pi_i)$ and $\mathbf{s}_q(\pi_j)$, respectively. Since different policies may use different scoring scales, directly comparing raw scores is unreliable. Therefore, Evo-B\&B first standardizes each score vector and converts it into a preference distribution:
\begin{equation}
\mathbf{p}_q(\pi)
=
\operatorname{softmax}
\left(
\frac{
\mathbf{s}_q(\pi)-\mu(\mathbf{s}_q(\pi))
}{
\sigma(\mathbf{s}_q(\pi))+\epsilon
}
\right)
\label{eq:probe_preference_distribution}
\end{equation}
where $\mu(\cdot)$ and $\sigma(\cdot)$ denote the mean and standard deviation of the score vector, respectively, and $\epsilon$ is a small constant used to avoid division by zero. This normalization preserves the relative preferences among candidates while reducing the effect of scale differences across policies.

For a decision probe $q$, Evo-B\&B first computes the top-choice consistency between two policies:
\begin{equation}
S_{\mathrm{top}}^q(\pi_i,\pi_j)
=
\begin{cases}
1, & \text{if } a_q(\pi_i)=a_q(\pi_j), \\
0, & \text{otherwise}
\end{cases}
\label{eq:probe_top_similarity}
\end{equation}
This term indicates whether the two policies select the same object from the same candidate set.

Evo-B\&B also compares the two policies' preference distributions over the entire candidate set. The distribution similarity is defined as
\begin{equation}
S_{\mathrm{dist}}^q(\pi_i,\pi_j)
=
1-
\frac{
\operatorname{JS}
\left(
\mathbf{p}_q(\pi_i),
\mathbf{p}_q(\pi_j)
\right)
}{
\log 2
}
\label{eq:probe_distribution_similarity}
\end{equation}
where $\operatorname{JS}(\cdot,\cdot)$ denotes the Jensen--Shannon divergence~\citep{nielsen2020generalization}. Since the Jensen--Shannon divergence lies in $[0,\log 2]$, $S_{\mathrm{dist}}^q$ is normalized to $[0,1]$, where a larger value indicates more similar preference distributions on the probe.

Finally, Evo-B\&B combines top-choice consistency and distribution similarity with equal weights to obtain the behavior similarity on a single decision probe:
\begin{equation}
S_q(\pi_i,\pi_j)
=
0.5 S_{\mathrm{top}}^q(\pi_i,\pi_j)
+
0.5 S_{\mathrm{dist}}^q(\pi_i,\pi_j)
\label{eq:probe_behavior_similarity_app}
\end{equation}

\subsection{Signature-level Similarity Aggregation}
\label{app:signature_level_aggregation}

After obtaining the similarity $S_q(\pi_i,\pi_j)$ on each matched decision probe, Evo-B\&B aggregates the similarities separately for branching probes and node-selection probes. Let $\mathcal{Q}_{\mathrm{branch}}(\pi_i,\pi_j)$ denote the set of matched branching probes shared by $\pi_i$ and $\pi_j$, and let $\mathcal{Q}_{\mathrm{node}}(\pi_i,\pi_j)$ denote the set of matched node-selection probes shared by the two policies. The branching behavior similarity is defined as
\begin{equation}
S_{\mathrm{branch}}(\pi_i,\pi_j)
=
\frac{1}{|\mathcal{Q}_{\mathrm{branch}}(\pi_i,\pi_j)|}
\sum_{q\in \mathcal{Q}_{\mathrm{branch}}(\pi_i,\pi_j)}
S_q(\pi_i,\pi_j)
\label{eq:branch_behavior_similarity}
\end{equation}
The node-selection behavior similarity is defined as
\begin{equation}
S_{\mathrm{node}}(\pi_i,\pi_j)
=
\frac{1}{|\mathcal{Q}_{\mathrm{node}}(\pi_i,\pi_j)|}
\sum_{q\in \mathcal{Q}_{\mathrm{node}}(\pi_i,\pi_j)}
S_q(\pi_i,\pi_j)
\label{eq:node_behavior_similarity}
\end{equation}

Finally, Evo-B\&B combines branching behavior similarity and node-selection behavior similarity with equal weights to obtain the overall behavior similarity between two joint policies:
\begin{equation}
S_{\mathrm{joint}}(\pi_i,\pi_j)
=
0.5S_{\mathrm{branch}}(\pi_i,\pi_j)
+
0.5S_{\mathrm{node}}(\pi_i,\pi_j)
\label{eq:joint_behavior_similarity_app}
\end{equation}
If no matched probes exist for one decision type, that component is excluded from the joint similarity calculation.

\section{Fitness}
\label{app:fitness}

Evo-B\&B evaluates each candidate joint policy $\pi$ based on its end-to-end SCIP solving performance on the training instance set $\mathcal{I}_{\mathrm{train}}$. For each instance $x\in\mathcal{I}_{\mathrm{train}}$, we record the solving time $T_\pi(x)$ and the number of explored nodes $N_\pi(x)$ obtained by policy $\pi$. Let $T_{\mathrm{base}}(x)$ and $N_{\mathrm{base}}(x)$ denote the corresponding results of default SCIP on instance $x$. Evo-B\&B first computes the geometric means of the solving-time ratio and node-count ratio over the training instances:
\begin{equation}
\bar{\rho}_T(\pi)
=
\mathrm{gmean}_{x\in\mathcal{I}_{\mathrm{train}}}
\frac{T_\pi(x)}{T_{\mathrm{base}}(x)},
\qquad
\bar{\rho}_N(\pi)
=
\mathrm{gmean}_{x\in\mathcal{I}_{\mathrm{train}}}
\frac{N_\pi(x)}{N_{\mathrm{base}}(x)}
\label{eq:fitness_ratios}
\end{equation}

Evo-B\&B then assigns equal importance to solving-time efficiency and node efficiency, and defines the end-to-end fitness of a candidate policy as
\begin{equation}
F(\pi)
=
\frac{100}{
\sqrt{
\bar{\rho}_T(\pi)\bar{\rho}_N(\pi)
}
}
\label{eq:end_to_end_fitness}
\end{equation}

A larger $F(\pi)$ indicates better overall solving efficiency relative to default SCIP. Policies that fail syntax, interface, or safety validation are regarded as invalid and are excluded from subsequent population updates. For executable policies that fail to solve all training instances within the time limit, Evo-B\&B applies an additional fitness penalty based on the solve rate and final optimality gap.

\section{Input Features}
\label{app:input_features}

Evo-B\&B uses features pre-extracted by the evaluator from the SCIP search process as inputs to the joint policy. 
The node-selection policy obtains information about the current open nodes and search state through 
\texttt{node\_features}, while the branching policy obtains information about the current LP branching candidates 
and their search context through \texttt{branch\_features}. Candidate policies are not allowed to directly call 
PySCIPOpt APIs to access additional internal solver states. Instead, all features are uniformly extracted by the 
fixed evaluation interface and passed to the policy functions, ensuring that different candidate policies are 
compared under the same information setting.

\paragraph{Global Features.}

Global features describe the overall search state of the current branch-and-bound (B\&B) process and are shared 
by both the branching and node-selection policies. These features include whether an incumbent solution exists, 
the primal-dual gap, the number of processed nodes, the number of open nodes, the maximum search depth, the number 
of feasible solutions found, and recent changes in the primal and dual bounds.

\paragraph{Node-selection Features.}

Node-selection features characterize the search state of open nodes and their relative positions within the 
current search frontier. They include node lower bound, estimate, depth, age, number of domain changes, bound and 
estimate improvements relative to the parent node or root node, as well as relative ranking information among the 
current open-node set.

\paragraph{Node-to-Branching Context.}

During branching-variable selection, each candidate variable is associated with the current focus node. To provide 
the branching policy with local search context, Evo-B\&B incorporates partial node-level features of the current 
focus node into the feature representation of each branching candidate.These features include the relative lower-bound and estimate positions of the current node among open nodes, depth 
ratio, bound improvement, and domain-change information. Since these features are identical for all candidate 
variables within the same branching callback, they mainly describe the search state of the current branching node 
and are combined with candidate-specific features for branching decisions.

\paragraph{Branching-candidate Features.}

Branching-candidate features describe the properties of candidate variables in the current LP relaxation. They 
include fractionality, LP solution value, objective coefficient, reduced cost, variable type, local and global 
domain widths, lock information, historical branching statistics, pseudocost values and their reliability proxies, 
as well as variable-constraint connectivity and coefficient statistics.

\paragraph{Branching-to-Node Coupling Features.}

Unlike Node-to-Branching Context, this category describes the historical branching decisions that generate the 
current node. To enable information interaction between branching and node-selection decisions, Evo-B\&B records 
the branching decision information that creates each child node and provides it as node features during subsequent 
node selection.These features include branching direction, branched-variable state, pseudocost and lock information, branching 
score margin, score entropy, and the predicted gain difference between the current child node and its sibling.

\paragraph{Feature Scaling and Normalization.}

To reduce numerical differences across instances and different search stages, Evo-B\&B provides stable scaled 
versions of continuous features.

Features with the \texttt{\_scaled} suffix use monotonic transformations that preserve the sign and relative 
ordering of the original values. Features with the \texttt{\_rank\_pct} suffix are normalized into the range 
$[0,1]$ to represent their relative ranks within the current candidate set.

To control the complexity of generated policies and improve interpretability, each candidate policy is restricted 
to using at most 15 feature fields.
\section{Benchmark Details}
\label{app:benchmark_details}
This section provides detailed configurations of the MILP benchmark instances used in our experiments.
All instances are generated with different random seeds and stored in the CIP format directly supported by SCIP.
The detailed instance generation configurations for each benchmark family are summarized in Table~\ref{tab:benchmark_config}.
\begin{table}[h]
\caption{Detailed configurations of the MILP benchmark datasets.}
\label{tab:benchmark_config}

\renewcommand{\arraystretch}{1.3}

\begin{center}
\begin{tabular}{ll}
\hline
\textbf{Benchmark} & \textbf{Configurations} \\
\hline

Set Covering &
\begin{tabular}[c]{@{}l@{}}
Easy: 500 rows, 1000 columns, density=0.05 \\
Medium: 1000 rows, 1000 columns, density=0.05 \\
Hard: 2000 rows, 1000 columns, density=0.05
\end{tabular}
\\[5pt]
\hline

Combinatorial Auctions &
\begin{tabular}[c]{@{}l@{}}
Easy: 100 items, 500 bids \\
Medium: 200 items, 1000 bids \\
Hard: 300 items, 1500 bids
\end{tabular}
\\[5pt]
\hline

Capacitated Facility Location &
\begin{tabular}[c]{@{}l@{}}
Easy: 100 customers, 100 facilities, ratio=5 \\
Medium: 200 customers, 100 facilities, ratio=5 \\
Hard: 400 customers, 100 facilities, ratio=5
\end{tabular}
\\[5pt]
\hline

Maximum Independent Set &
\begin{tabular}[c]{@{}l@{}}
Easy: 750 nodes, affinity=4 \\
Medium: 1000 nodes, affinity=4 \\
Hard: 1500 nodes, affinity=4
\end{tabular}
\\[5pt]
\hline

\end{tabular}
\end{center}
\end{table}
\section{Additional Training Details}
\label{app:training_details}

This section provides detailed configurations of the LLM, multi-island evolution hyperparameters, behavior probe construction, and parameter tuning used in Evo-B\&B.

\paragraph{LLM Configuration.}
All policies are generated using \texttt{qwen3-max-2026-01-23}. Throughout the evolutionary process, the temperature and top-$p$ are set to $0.7$ and $0.95$, respectively, and the maximum output length for each generation is set to $8192$ tokens.
\paragraph{Multi-Island Evolution.}
Evo-B\&B maintains $4$ parallel islands, with a maximum population capacity of $8$ for each island. At each evolution round, each island generates $1$ offspring and constructs a parent set containing up to $3$ candidate policies for mutation or crossover.

For each benchmark, the evolutionary search runs for at most $150$ rounds with a wall-clock budget of $10$ hours, and terminates when either limit is reached. Since the time limit is checked at iteration or evaluation-batch boundaries, ongoing SCIP evaluations are allowed to finish, which may result in the recorded runtime slightly exceeding $10$ hours.

Cross-island information migration is triggered every $10$ rounds. For each target island, policies with high quality and behavioral differences from the local population are selected from other islands, and $2$ design ideas are extracted to guide subsequent LLM generations. An island reset is triggered when no performance improvement is observed for $8$ consecutive rounds, preserving the local elite and injecting the current global best to redirect the search.
\paragraph{Solver Configuration.}
All experiments are conducted using SCIP 10.0.1, PySCIPOpt 6.1.0, and Python 3.10.11. To ensure that different strategies are compared under a consistent solving environment, we use the same SCIP configuration for all methods. Cutting-plane separation is disabled at non-root nodes, while the default separation process at the root node is retained. We also retain the initial presolving stage before the search starts, but disable presolving restarts during the Branch-and-Bound process.

\section{Evolution of the Discovered Joint Policy}
\label{app:policy_evolution}

\begin{figure}[htb]
    \centering
    \includegraphics[width=\linewidth]{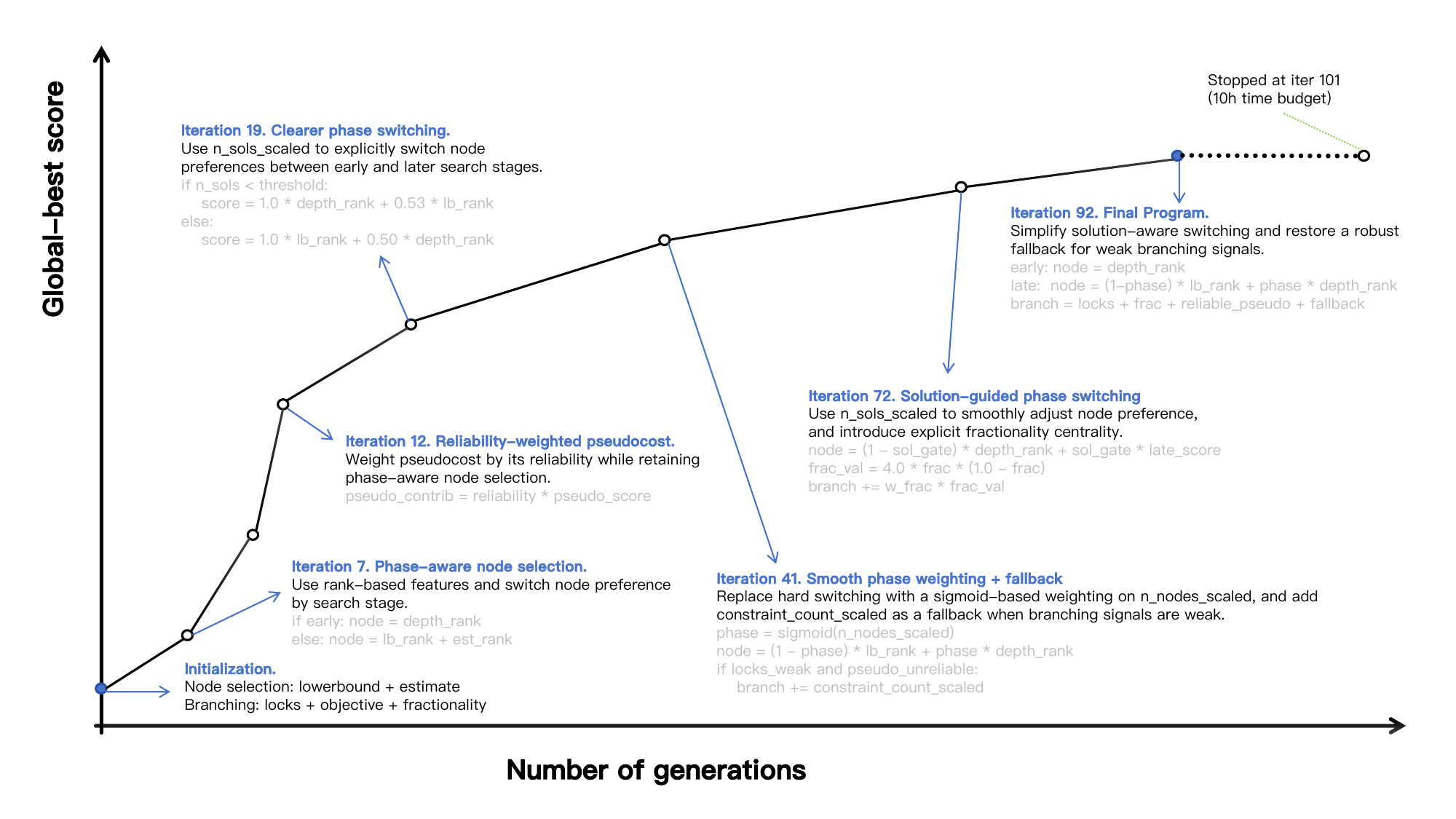}
    \caption{Evolution of the joint policy discovered by Evo-B\&B on Indset. The curve tracks the global-best score and highlights representative policy updates.}
    \label{fig:policy_evolution}
\end{figure}

This section analyzes the evolution of the joint policy discovered by Evo-B\&B on Indset. Figure~\ref{fig:policy_evolution} shows the trajectory of the global-best score during training. As evolution progresses, the node-selection policy gradually evolves from a static feature combination into a phase-aware adaptive rule, while the branching policy progressively incorporates pseudocost reliability, fractionality centrality, and fallback signals for cases where the primary branching signals are weak.

\begin{itemize}

\item \textbf{Initialization.}
The initial global-best policy adopts relatively direct scoring rules. Node selection jointly considers the lower bound and estimate of each node, while branching combines locks, the objective coefficient, and fractionality. The policy mainly relies on raw solver features and does not yet adapt its node-selection behavior to different stages of the B\&B search.

\item \textbf{Iteration 7: Phase-aware node selection.}
The node-selection policy begins to use rank-based features and introduces a search-phase switching mechanism. During the early stage, it places greater emphasis on the relative depth ranking \texttt{depth\_rank}, and gradually shifts toward the relative rankings of the node lower bound and estimate, \texttt{lowerbound\_rank} and \texttt{estimate\_rank}. This change allows node selection to adapt its preference according to the current search stage rather than relying on a fixed global rule.

\item \textbf{Iteration 12: Reliability-weighted pseudocost.}
While retaining phase-aware node selection, the branching policy further incorporates pseudocost reliability. The contribution of pseudocost to the branching score is weighted by its reliability, following the form
\texttt{reliability $\times$ pseudo\_score},
so that less reliable pseudocost information has a smaller influence on the final branching decision.

\item \textbf{Iteration 19: Clearer phase switching.}
The node-selection policy further distinguishes the decision rules used at different search stages. It uses \texttt{n\_sols\_scaled}, obtained by applying a $\log(1+x)$ transformation to the number of feasible solutions found so far, to characterize the current search stage. During the early stage, the policy places more emphasis on the relative node-depth ranking \texttt{depth\_rank}; as the search progresses, it increases the importance of the relative lower-bound ranking \texttt{lowerbound\_rank}. The branching policy continues to use reliability-weighted pseudocost, reducing the influence of unreliable pseudocost information on branching scores.

\item \textbf{Iteration 41: Smooth phase weighting and fallback.}
The phase-switching mechanism further evolves into a smooth adjustment based on the number of explored search nodes. The log-compressed node count \texttt{n\_nodes\_scaled} is used as a signal of search progress, and a sigmoid function continuously adjusts the weights of \texttt{lowerbound\_rank} and \texttt{depth\_rank}.

\item \textbf{Iteration 72: Solution-guided phase switching.}
The node-selection policy further exploits the number of feasible solutions found during the current search. Based on \texttt{n\_sols\_scaled}, it constructs a smooth gating coefficient that continuously switches between an early-stage rule dominated by the relative node-depth ranking \texttt{depth\_rank} and a later-stage rule combining lower-bound and depth information. Meanwhile, the branching policy rewrites the fractionality term as $4f(1-f)$, assigning higher scores to candidate variables whose LP relaxation values are closer to $0.5$, thereby more explicitly favoring highly fractional variables.

\item \textbf{Iteration 92: Final program.}
The final policy consolidates and simplifies the effective mechanisms discovered in earlier stages. Node selection directly uses \texttt{depth\_rank} during the early search stage and subsequently applies a smooth weighting between \texttt{lowerbound\_rank} and \texttt{depth\_rank} according to search progress. The branching policy retains fractionality, reliability-weighted pseudocost, and locks, while using \texttt{constraint\_count\_scaled} as a fallback when the primary branching signals are weak. This policy becomes the final global-best policy, and the evolutionary search terminates at iteration 101 after reaching the $10$-hour time budget.

\end{itemize}

Overall, the evolutionary process on Indset does not simply increase the number of features or the complexity of the policy. Instead, it exhibits a gradual process of refinement and simplification. Node selection evolves from an initial static scoring rule into an adaptive policy that adjusts its behavior according to the search stage, while branching progressively incorporates reliability-weighted pseudocost, fractionality, and fallback signals. As evolution proceeds, some of the more complex mechanisms are further simplified and consolidated into the final global-best policy, resulting in a relatively compact joint branching and node-selection strategy.

\section{Computational Cost}
\label{app:computational_cost}
We further report the actual evolutionary cost of Evo-B\&B on the four benchmarks, including the number of evolution rounds, wall-clock time, LLM calls, token consumption, and instance-level SCIP evaluations during training. Token counts are reported in millions (M). Solver Calls denote instance-level SCIP solving runs, and each candidate policy is evaluated on $10$ training instances. Detailed statistics are provided in Table~\ref{tab:computational_cost}.

\begin{table}[H]
\centering
\caption{
Computational cost of Evo-B\&B during evolutionary search.
Token counts are reported in millions (M), and Solver Calls denote instance-level SCIP evaluations.
}
\label{tab:computational_cost}

\small
\setlength{\tabcolsep}{4.2pt}
\renewcommand{\arraystretch}{1.0}

\begin{tabular*}{\linewidth}
{@{\extracolsep{\fill}}lrrrrrrr@{}}
\toprule
Problem
& Iter.
& Time (h)
& LLM Calls
& Input (M)
& Output (M)
& Total (M)
& Solver Calls \\
\midrule

Setcover
& 150/150
& 9.42
& 600
& 3.97
& 0.87
& 4.84
& 11,230 \\

Cauctions
& 150/150
& 3.15
& 600
& 3.34
& 0.66
& 4.00
& 7,340 \\

Indset
& 101/150
& 10.02
& 404
& 2.24
& 0.39
& 2.64
& 6,180 \\

Facilities
& 100/150
& 10.25
& 400
& 2.28
& 0.46
& 2.74
& 5,840 \\

\midrule

\textbf{Total}
& 501/600
& 32.83
& 2,004
& 11.83
& 2.39
& 14.22
& 30,590 \\

\bottomrule
\end{tabular*}
\end{table}

For each benchmark, the evolutionary search terminates when either the maximum number of evolution rounds (150) or the 10-hour wall-clock budget is reached, whichever comes first. Setcover and Cauctions complete all 150 rounds, whereas Indset and Facilities reach the time budget at rounds 101 and 100, respectively. Since the wall-clock limit is checked only at iteration or evaluation-batch boundaries, SCIP evaluations that have already started are allowed to finish, which can result in the recorded runtime slightly exceeding 10 hours. In total, the four evolutionary runs complete 501 rounds, invoke the LLM 2,004 times, consume approximately 14.22M tokens, and perform 30,590 instance-level SCIP evaluations.

\section{Prompt Engineering Details}
\label{app:prompt_design}

Evo-B\&B adopts a structured prompting scheme consisting of a System Prompt
and a User Prompt. The System Prompt remains fixed throughout evolution and
defines the role of the LLM, the optimization objective, the executable
interface, safety constraints, and the required output format. The User Prompt
is dynamically constructed according to the selected evolution operator and
provides parent policies, performance feedback, migration information, and
operator-specific transformation instructions.

\subsection{System Prompt}
\label{app:shared_system_prompt}

All evolution operators share the following fixed System Prompt. The search
direction of each operator is controlled by its corresponding User Prompt.
The complete input feature definitions are provided in
Appendix~\ref{app:input_features}.

\begin{promptbox}{System Prompt}

You are an expert in MILP solver design with PySCIPOpt. Your task is to jointly
improve the node-selection scoring function and the branching-variable scoring
function.

\textbf{Optimization Objective.}

Candidate joint policies are evaluated according to their end-to-end SCIP
solving performance on MILP instances. The fitness jointly considers the
solving-time ratio and node-count ratio relative to default SCIP, with larger
fitness indicating better performance. Subject to correct solving behavior,
the objective is to reduce both solving time and the number of B\&B search
nodes.

The current stage focuses on structural evolution. The search should primarily
explore scoring structures, feature combinations, and decision mechanisms
rather than merely modifying numerical parameter values. Continuous parameters
are further optimized in a separate parameter-optimization stage.

\textbf{Joint Policy Interface.}

The generated joint policy must contain a tunable parameter list
\texttt{PARAMS}, the corresponding parameter bounds \texttt{BOUNDS}, a
node-selection scoring function, and a branching-variable scoring function.
The node-selection scoring function returns a finite numerical score for an
open node, where a larger score indicates higher selection priority. The
branching-variable scoring function returns scores for the current LP branching
candidates, where a larger score indicates higher branching priority.

\textbf{Feature Usage.}

Candidate policies may only use the pre-extracted B\&B features provided
through the fixed scoring interface and must not directly query additional
solver information from PySCIPOpt objects. The complete feature definitions
are provided in Appendix~\ref{app:input_features}.

\textbf{Safety Constraints.}

Generated policies must not modify the internal SCIP solving state or directly
call PySCIPOpt APIs to obtain additional features. All numerical computations
must remain stable, and the scoring functions must return only finite values
without producing \texttt{NaN} or infinite values.

\textbf{Structural Evolution Requirements.}

Candidate policies must introduce meaningful structural changes rather than
merely modifying parameter values, parameter ordering, variable names, or
comments. Structural changes may include feature normalization or nonlinear
transformations, search-state-dependent gating or piecewise scoring,
lexicographic priorities, reliability-aware scoring, and coupling mechanisms
between branching and node selection. To keep the generated policies compact
and suitable for subsequent parameter optimization, each policy may use at
most $15$ feature fields and $15$ tunable parameters.

\textbf{Output Format.}

The output must strictly contain the following three sections:
\texttt{[Thought]}, \texttt{[KEY PARAMETERS]}, and \texttt{[Code]}.

\texttt{[Thought]} describes the search intuition, the main structural
modification, and its expected effect.

\texttt{[KEY PARAMETERS]} lists the initial values of the tunable parameters,
their roles, and the corresponding parameter ranges, together with the
rationale for these ranges.

\texttt{[Code]} provides the complete executable implementation of the joint
policy.

\end{promptbox}

\subsection{User Prompt}
\label{app:user_prompts}

Following the operator-based evolutionary prompting paradigm in
EoH~\citep{liu2024evolution}, Evo-B\&B instantiates different structural
transformations through operator-specific User Prompts.The User Prompt of Evo-B\&B consists of operator-specific instructions and
dynamic search context. For mutation operators, the User Prompt provides the
code, design rationale, key parameters, and solving feedback of the primary
parent. For crossover operators, the corresponding information of the primary
parent and up to two auxiliary parents is provided. The operator-specific
instructions determine the policy transformation performed in the current
generation.

Cross-island migration information is maintained in the migration memory of
each target island. Whenever periodic migration is triggered, Evo-B\&B ranks
candidate policies from the other islands according to the migration utility
and selects the top $2$ external policies for each target island as information
sources. These policies are not directly copied into the target population.
Instead, their source island, source-policy fitness, solving-time ratio,
node-count ratio, and design rationale are extracted and stored in the
migration memory. When constructing the User Prompt for regular candidate
generation, Evo-B\&B first sorts the stored migration information from the most
recent to the oldest migration round. Records from the same migration round
are further ranked by the fitness of their source policies. The top $3$
records are provided as auxiliary context, while the LLM is instructed to
incorporate at most $1$ migrated idea into a single generated policy.

When an island reset is triggered, the fusion-reset operator $r_1$ uses a
dedicated reset prompt. The prompt provides the complete code and solving
feedback of the local best policy and the global best policy, together with
summary information about the external policies directly introduced into the
reset population, including their source islands, fitness values, solving-time
ratios, and node-count ratios. In addition, the reset prompt retrieves the
migration information accumulated during previous periodic migrations. Using
the same ranking rule, at most the top $4$ migration records are retained as
auxiliary context.

\paragraph{Structural Mutation ($m_1$).}
The $m_1$ operator performs a single-parent structural mutation by modifying
one core component of the joint policy while keeping the other component
largely unchanged.

\begin{promptbox}{User Prompt for Structural Mutation ($m_1$)}

\textbf{Parent Strategy}

Parent thought: \{\texttt{parent\_thought}\}

Parent key parameters: \{\texttt{parent\_key\_parameters}\}

Parent important results: \{\texttt{parent\_metrics}\}

Parent code: \{\texttt{parent\_code}\}

\textbf{Task}

Perform a single-parent structural mutation on the given parent policy by
modifying one core structure in either the node-selection scoring function or
the branching-variable scoring function.

\textbf{Requirements}

Modify only one of the two scoring functions and keep the other algebraically
close to the parent. Prefer one clear and interpretable structural change, such
as search-phase switching, feature normalization, reliability modeling,
lexicographic priority, fallback mechanisms, depth--bound--estimate trade-offs,
or interactions among pseudocost, fractionality, lock, and constraint
information.

Do not merely modify numerical parameter values. Migration information may
guide the local structural mutation, but at most one migrated idea may be used,
and the complete mechanism or code of another strategy must not be directly
copied. The goal is to produce an interpretable local variant with a clear
structural difference from its parent.

\textbf{Migrated Ideas}

\{\texttt{migrated\_ideas}\}

\end{promptbox}

\paragraph{Scoring Logic Refinement ($m_2$).}
The $m_2$ operator performs single-parent scoring-logic refinement by adjusting
a local scoring mechanism while keeping the overall structure of the parent
policy largely unchanged, thereby improving discrimination among candidate
nodes or branching variables.

\begin{promptbox}{User Prompt for Scoring Logic Refinement ($m_2$)}

\textbf{Parent Strategy}

Parent thought: \{\texttt{parent\_thought}\}

Parent key parameters: \{\texttt{parent\_key\_parameters}\}

Parent important results: \{\texttt{parent\_metrics}\}

Parent code: \{\texttt{parent\_code}\}

\textbf{Task}

Refine the scoring logic of the parent policy while keeping its overall
structure largely unchanged, so as to better distinguish candidate nodes or
branching variables.

\textbf{Requirements}

Preserve the main structure of the parent policy and introduce only one local
enhancement. Prefer refinements to gate calibration, fallback ordering,
tie-breaking priority, feature normalization, scale compression, or weight
interaction. A weak secondary feature may be replaced, but the overall strategy
shape should remain unchanged.

The strategy complexity may be preserved or slightly increased, but too many
new tunable parameters should not be introduced. Do not perform crossover.
Migration information may guide the local refinement, but at most one migrated
idea may be used, and the complete mechanism or code of another strategy must
not be directly copied. The goal is to improve the stability and fine-grained
control of the scoring logic while preserving the main parent structure.

\textbf{Migrated Ideas}

\{\texttt{migrated\_ideas}\}

\end{promptbox}

\paragraph{Regularization Mutation ($m_3$).}
The $m_3$ operator performs a single-parent regularization mutation that
simplifies redundant or unstable scoring logic while preserving the core
search preference of the parent policy.

\begin{promptbox}{User Prompt for Regularization Mutation ($m_3$)}

\textbf{Parent Strategy}

Parent thought: \{\texttt{parent\_thought}\}

Parent key parameters: \{\texttt{parent\_key\_parameters}\}

Parent important results: \{\texttt{parent\_metrics}\}

Parent code: \{\texttt{parent\_code}\}

\textbf{Task}

Perform a single-parent regularization mutation on the given parent policy,
simplifying or stabilizing the strategy while preserving its core search
preference.

\textbf{Requirements}

Remove redundant, brittle, overly complex, or difficult-to-interpret scoring
logic, and reduce unnecessary parameters, conditional branches, or repeated
features. Preserve the most central and stable search mechanism of the parent,
and replace complex scoring expressions with simpler normalization or
tie-breaking rules when appropriate.

Do not introduce new feature fields, tunable parameters, gates, or phase
switches. Migration information may be used to guide which mechanisms should
be retained or removed, but at most one migrated idea may be used, and no
mechanism or code from another strategy may be introduced. The goal is to
reduce strategy complexity while improving robustness and interpretability
without changing the main search behavior.

\textbf{Migrated Ideas}

\{\texttt{migrated\_ideas}\}

\end{promptbox}

\paragraph{Explorative Crossover ($e_1$).}
The $e_1$ operator performs multi-parent explorative crossover by recombining
node-selection and branching mechanisms from different parents to explore new
joint-policy structures with different component combinations.

\begin{promptbox}{User Prompt for Explorative Crossover ($e_1$)}

\textbf{Crossover Parent Set}

\{\texttt{crossover\_parents}\}

\textbf{Task}

Perform an explorative node--branch crossover using at least two equal-rank
parent policies. Recombine node-selection and branching mechanisms from
different parents to generate a new joint policy while preserving policy
validity and solving reliability.

\textbf{Requirements}

Treat all provided parents as equal design sources rather than preselecting one
parent as the complete policy backbone. The node-selection mechanism should be
inherited primarily from one parent and the branching mechanism primarily from
another, or the two components should be explicitly recombined across different
parents.

Beyond parent-mechanism recombination, at most one additional local mechanism,
such as a gate, fallback rule, or tie-breaking rule, may be introduced.
In \texttt{[Thought]}, explicitly state which parents provide the
node-selection and branching mechanisms.

Do not directly splice unrelated code fragments. Instead, reorganize useful
mechanisms from different parents into a coherent and executable joint policy.
The goal is behaviorally meaningful cross-component recombination rather than
a single-parent mutation.

\textbf{Migrated Ideas}

\{\texttt{migrated\_ideas}\}

\end{promptbox}

\paragraph{Backbone Extraction Crossover ($e_2$).}
The $e_2$ operator performs multi-parent backbone extraction crossover by
identifying repeatedly successful mechanisms shared across high-quality
parents and making local adjustments around the resulting common backbone.

\begin{promptbox}{User Prompt for Backbone Extraction Crossover ($e_2$)}

\textbf{Crossover Parent Set}

\{\texttt{crossover\_parents}\}

\textbf{Task}

Perform backbone extraction crossover using $2$--$3$ equal-rank,
high-quality parent policies. Identify common or repeatedly successful
mechanisms across the parents and synthesize a new joint policy that preserves
the shared backbone while introducing a small number of justified variations.

\textbf{Requirements}

Treat all provided parents as evidence for backbone extraction rather than
mutating a single parent. In \texttt{[Thought]}, explicitly describe the
identified common backbone, such as best-estimate or best-bound node-selection
pressure, reliability-aware pseudocost branching, fractionality preference,
incumbent-aware phase switching, or stable tie-breaking mechanisms.

Unless one component is clearly weak, preserve the shared backbone in both
node selection and branching. Modify only one or two local design points
related to normalization, tie-breaking, pseudocost reliability, phase gates,
fallback mechanisms, or secondary features.

Do not destroy the shared backbone or collapse into a direct copy of a single
parent. The goal is to generate a stable recombination based on mechanisms
shared by multiple high-quality strategies.

\textbf{Migrated Ideas}

\{\texttt{migrated\_ideas}\}

\end{promptbox}

\paragraph{Fusion Reset ($r_1$).}
The $r_1$ operator is activated during island reset and combines the local best
policy, global best policy, and cross-island migration information to generate
a new joint policy for a stagnated island.

\begin{promptbox}{User Prompt for Fusion Reset ($r_1$)}

\textbf{Local Best Policy}

Local metrics: \{\texttt{parent\_metrics}\}

Local code: \{\texttt{parent\_code}\}

\textbf{Global Best Policy}

Global metrics: \{\texttt{best\_metrics}\}

Global code: \{\texttt{best\_code}\}

\textbf{Migrated Ideas}

\{\texttt{migrated\_ideas}\}

\textbf{Reset Context}

\{\texttt{fusion\_context}\}

\textbf{Task}

Generate a fusion-reset candidate for the stagnated island by jointly using
the local best policy, the global best policy, and cross-island migration
information.

\textbf{Requirements}

Preserve the most reliable backbone of the global best policy, while
incorporating at most one specialized mechanism from the local best policy and
at most one migrated idea.

Do not directly copy any existing policy. Instead, synthesize useful mechanisms
from different sources into a coherent and executable joint policy. Maintain
policy validity and solving reliability while jointly improving solving time
and node count, with preference for reducing the node-count ratio without
substantially degrading the solving-time ratio.

\end{promptbox}

\section{Discovered Joint Policies}
\label{app:discovered_policies}

\subsection{Discovered Joint Policy on Setcover}
\label{app:setcover_policy}

\begin{lstlisting}[style=compactpython]
import math
import numpy as np
from branch_node_evolve.strategy_helpers import safe_float, frac_center

PARAMS = [0.4100749645335086, -1.2271610375629374, -0.22331832051812983, 0.15, 0.18727388668602052, 1.3763385870332128, 0.2562352276743422, 0.46886715543492813, 0.36712579655493466]
BOUNDS = [[0.0, 1.0], [-2.0, 0.0], [-2.0, 0.0], [0.0, 0.5], [0.0, 1.0], [0.0, 2.0], [-1.0, 1.0], [0.1, 1.0], [0.0, 1.0]]

def node_score(model, node_features, params):
    depth_rank = safe_float(node_features.get('depth_rank_pct'))
    lowerbound_rank = safe_float(node_features.get('lowerbound_rank_pct'))
    estimate_rank = safe_float(node_features.get('estimate_rank_pct'))
    n_nodes_scaled = safe_float(node_features.get('n_nodes_scaled'))
    gap_scaled = safe_float(node_features.get('gap_scaled'))

    early_phase_weight = max(0.0, 1.0 - n_nodes_scaled / max(params[4], 1e-06))
    late_phase_weight = 1.0 - early_phase_weight

    depth_contribution = params[0] * early_phase_weight * depth_rank
    bound_contribution = params[1] * late_phase_weight * lowerbound_rank
    estimate_contribution = params[2] * late_phase_weight * estimate_rank
    score = depth_contribution + bound_contribution + estimate_contribution

    has_origin = safe_float(node_features.get('has_origin_annotation'), 0.0) > 0.5
    if has_origin:
        dir_adv = safe_float(node_features.get('direction_advantage_scaled'), 0.0)
        score += params[3] * dir_adv
    return score

def branch_score(model, branch_features, params):
    if not branch_features:
        return np.array([], dtype=float)

    scores = np.zeros(len(branch_features), dtype=float)
    for idx, feat in enumerate(branch_features):
        frac_centrality = safe_float(feat.get('frac_centrality'), 0.0)
        locks = safe_float(feat.get('locks_sum_scaled'), 0.0)
        obj = abs(safe_float(feat.get('obj_scaled'), 0.0))

        base_score = params[5] * frac_centrality + params[6] * locks + params[5] * math.log1p(obj)

        pseudo_score = safe_float(feat.get('pseudo_score_scaled'), 0.0)
        local_reliability = safe_float(feat.get('pseudo_reliability_proxy_scaled'), 0.0)
        global_reliability = safe_float(feat.get('origin_pseudo_reliability_proxy_scaled'), 0.0)

        mixed_reliability = params[8] * local_reliability + (1.0 - params[8]) * global_reliability
        effective_reliability = max(0.0, min(1.0, mixed_reliability))
        blend_weight = effective_reliability ** params[7]

        scores[idx] = base_score + blend_weight * params[6] * pseudo_score

    return scores
\end{lstlisting}

\subsection{Discovered Joint Policy on Indset}
\label{app:indset_policy}

\begin{lstlisting}[style=compactpython]
import math
import numpy as np
from branch_node_evolve.strategy_helpers import safe_float, frac_center

PARAMS = [0.12183626369988391, 0.7458895514456119, 1.164073236041003, 0.8509613531742889, 0.761453248822705, 0.08372305891153102, 0.1938708785152441]
BOUNDS = [[-2.0, 2.0], [-2.0, 2.0], [0.0, 2.0], [0.0, 2.0], [0.0, 2.0], [0.0, 1.0], [0.0, 1.0]]

def node_score(model, node_features, params):
    lb_rank = safe_float(node_features.get('lowerbound_rank_pct'))
    depth_rank = safe_float(node_features.get('depth_rank_pct'))
    n_nodes = safe_float(node_features.get('n_nodes_scaled'))
    n_sols = safe_float(node_features.get('n_sols_scaled'))

    if n_sols < params[5]:
        return depth_rank

    phase_weight = 1.0 / (1.0 + math.exp(-8.0 * (n_nodes - 0.5)))
    score = (1.0 - phase_weight) * (params[0] * lb_rank) + phase_weight * (params[1] * depth_rank)
    return score

def branch_score(model, branch_features, params):
    scores = np.zeros(len(branch_features), dtype=float)

    for idx, feat in enumerate(branch_features):
        locks = safe_float(feat.get('locks_sum_scaled'))
        frac = safe_float(feat.get('frac'))
        frac_val = 4.0 * frac * (1.0 - frac) if 0.0 <= frac <= 1.0 else 0.0

        pseudo_score = safe_float(feat.get('pseudo_score_scaled'))
        reliability = safe_float(feat.get('pseudo_reliability_proxy_scaled'))
        constraint_count = safe_float(feat.get('constraint_count_scaled'))

        pseudo_contrib = params[4] * reliability * pseudo_score
        lock_weak = locks < 0.3
        pseudo_unreliable = reliability < 0.4
        fallback = params[6] * constraint_count if lock_weak and pseudo_unreliable else 0.0

        scores[idx] = params[2] * locks + params[3] * frac_val + pseudo_contrib + fallback

    return scores
\end{lstlisting}
\subsection{Discovered Joint Policy on Facilities}
\label{app:facilities_policy}

\begin{lstlisting}[style=compactpython]
import math
import numpy as np
from branch_node_evolve.strategy_helpers import safe_float, frac_center

PARAMS = [0.9443693997717498, -0.315079119462618, -0.2, 2.7641272567834827, 0.6845174200975173, 0.0284815317535618, 0.17166054015606155, 0.3996302564147792, 0.0005, 0.5710087669921102]
BOUNDS = [[0.0, 2.0], [-3.0, 3.0], [-3.0, 3.0], [-4.0, 5.0], [-2.0, 3.0], [-1.0, 1.0], [-2.0, 3.0], [0.0, 1.0], [0.0, 1.0], [0.0, 1.0]]

def node_score(model, node_features, params):
    depth_rank = safe_float(node_features.get('depth_rank_pct'))
    lowerbound_rank = safe_float(node_features.get('lowerbound_rank_pct'))
    estimate_rank = safe_float(node_features.get('estimate_rank_pct'))
    lowerbound_rel_best = safe_float(node_features.get('lowerbound_rel_best_scaled'), default=0.0)
    estimate_rel_best = safe_float(node_features.get('estimate_rel_best_scaled'), default=0.0)
    has_incumbent = bool(node_features.get('has_incumbent', False))
    gap = safe_float(node_features.get('gap_scaled'), default=1.0)
    n_nodes_norm = safe_float(node_features.get('n_nodes_scaled'), default=0.0)

    phase_threshold = params[8]

    if n_nodes_norm < phase_threshold:
        if has_incumbent:
            blend_weight = params[9]
            score = (1.0 - blend_weight) * lowerbound_rank + blend_weight * estimate_rank
        else:
            score = lowerbound_rank
    else:
        bound_quality = max(lowerbound_rel_best, estimate_rel_best) if has_incumbent else lowerbound_rel_best
        gated_depth = depth_rank * (1.0 + bound_quality)
        bound_weight = params[1] * gap
        estimate_weight = params[2] * (1.0 - gap) if has_incumbent else 0.0
        score = params[0] * gated_depth + bound_weight * lowerbound_rank + estimate_weight * estimate_rank

    return score

def branch_score(model, branch_features, params):
    scores = np.zeros(len(branch_features), dtype=float)

    for idx, feat in enumerate(branch_features):
        frac_cent = safe_float(feat.get('frac_centrality'))
        frac_cent_sq = frac_cent ** 2
        locks = safe_float(feat.get('locks_sum_scaled'))
        obj = safe_float(feat.get('obj_scaled'))

        base_score = params[3] * frac_cent_sq + params[4] * locks + params[5] * obj

        pseudo_score = safe_float(feat.get('pseudo_score_scaled'))
        reliable_proxy = safe_float(feat.get('pseudo_reliable_proxy'))
        reliability_threshold = params[7]

        if reliable_proxy >= reliability_threshold:
            base_score += params[6] * pseudo_score

        scores[idx] = base_score

    return scores
\end{lstlisting}

\subsection{Discovered Joint Policy on Cauctions}
\label{app:cauctions_policy}

\begin{lstlisting}[style=compactpython]
import math
import numpy as np
from branch_node_evolve.strategy_helpers import safe_float, frac_center

PARAMS = [1.4883681987646005, -0.12043203281358458, 0.1, 1.9825701167007799, 0.7357557817653823, 0.25, 0.04189328119249113]
BOUNDS = [[0.5, 3.0], [-1.0, 1.0], [-0.5, 0.5], [1.0, 3.0], [0.1, 2.0], [0.0, 0.5], [0.01, 0.2]]

def node_score(model, node_features, params):
    has_incumbent = bool(node_features.get('has_incumbent', False))
    gap_scaled = safe_float(node_features.get('gap_scaled', 1.0))
    estimate_rank = safe_float(node_features.get('estimate_rank_pct', 0.0))
    lowerbound_rank = safe_float(node_features.get('lowerbound_rank_pct', 0.0))
    depth_rank = safe_float(node_features.get('depth_rank_pct', 0.0))

    score = params[0] * estimate_rank + params[1] * lowerbound_rank + params[2] * depth_rank

    if has_incumbent and gap_scaled < params[6]:
        score += 0.5 * estimate_rank

    return score

def branch_score(model, branch_features, params):
    if not branch_features:
        return np.array([], dtype=float)

    scores = np.zeros(len(branch_features), dtype=float)

    for idx, feat in enumerate(branch_features):
        frac_cent = safe_float(feat.get('frac_centrality', 0.0))
        locks_sum = abs(safe_float(feat.get('locks_sum', 0.0)))
        obj = abs(safe_float(feat.get('obj', 0.0)))

        score = params[3] * frac_cent \
                + params[4] * math.log1p(locks_sum) \
                + params[5] * math.log1p(obj)

        scores[idx] = score

    return scores
\end{lstlisting}

\end{document}